\documentclass[10pt,journal,compsoc]{IEEEtran}

\usepackage{graphicx} % Required for inserting images
\usepackage{amssymb}
\usepackage{amsmath}
\usepackage{algorithm}
\usepackage{algorithmic}
\usepackage{cite}
\usepackage{multirow}
\usepackage{verbatim}
\usepackage{booktabs}
\usepackage{epsfig}
\usepackage{subcaption}
\usepackage{color}
\usepackage{amsthm}
\usepackage{mathrsfs}
\usepackage{amsfonts}
\usepackage{booktabs}
\usepackage{float}

\newcommand{\bA}{\mathbf{A}}

\newcommand{\bc}{\mathbf{c}}

\newcommand{\bD}{\mathbf{D}}
\newcommand{\bW}{\mathbf{W}}
\newcommand{\bC}{\mathbf{C}}
\newcommand{\bZ}{\mathbf{Z}}

\newcommand{\bX}{\mathbf{X}}

\newcommand{\bb}{\mathbf{b}}
\newcommand{\bx}{\mathbf{x}}
\newcommand{\bI}{\mathbf{I}}
\newcommand{\bR}{\mathbf{R}}
\newcommand{\bPi}{\boldsymbol{\Pi}}

\newcommand{\bu}{\mathbf{u}}

\newcommand{\tbmu}{\tilde{\boldsymbol{\mu}}}
\newcommand{\tsigma}{\tilde{\sigma}}
\newcommand{\bz}{\mathbf{z}}

\newcommand{\bbR}{\mathbb{R}}
\newcommand{\bbE}{\mathbb{E}}

\newcommand{\calZ}{\boldsymbol{\mathcal{Z}}}

\newcommand{\calN}{\mathcal{N}}
\newcommand{\calX}{\mathcal{X}}
\newcommand{\calB}{\mathcal{B}}

\newcommand{\calE}{\mathcal{E}}

\newcommand{\calU}{\mathcal{U}}
\newcommand{\bcalB}{\boldsymbol{\calB}}

\newcommand{\bDelta}{\boldsymbol{\Delta}}
\newcommand{\bcalX}{\boldsymbol{\calX}}
\newcommand{\bcalZ}{\boldsymbol{\calZ}}
\newcommand{\bcalE}{\boldsymbol{\calE}}
\newcommand{\bcalU}{\boldsymbol{\calU}}

\newcommand{\bgamma}{\boldsymbol{\gamma}}
\newcommand{\bSigma}{\boldsymbol{\Sigma}}
\newcommand{\bGamma}{\boldsymbol{\Gamma}}

\newtheorem{assumption}{Assumption}
\newtheorem{lemma}{Lemma}

\newtheorem{theorem}{Theorem}

\newtheorem{remark}{Remark}
\newtheorem{definition}{Definition}

\DeclareMathOperator{\cov}{Cov}

\ifCLASSINFOpdf
\else
\fi
\begin{document}
\title{Tensor State Space-based Dynamic Multilayer Network Modeling}

\author{Tian~Lan\thanks{Tian Lan is with the Department of Industrial Engineering, Tsinghua University, Beijing, China (e-mail: lant23@mails.tsinghua.edu.cn).}, 
        Jie~Guo\thanks{Jie Guo is with the Department of Industrial Engineering, Tsinghua University, Beijing, China (e-mail: guojie19@mails.tsinghua.edu.cn).}, 
        and Chen~Zhang\thanks{Chen Zhang is with the Department of Industrial Engineering, Tsinghua University, Beijing, China (e-mail: zhangchen01@tsinghua.edu.cn).}
}
% \author{Anonymous authors}


\markboth{Journal of \LaTeX\ Class Files,~Vol.~14, No.~8, August~2015}%
{Shell \MakeLowercase{\textit{et al.}}: Bare Demo of IEEEtran.cls for IEEE Journals}



\maketitle

\begin{abstract}
Understanding the complex interactions within dynamic multilayer networks is critical for advancements in various scientific domains. Existing models often fail to capture such networks' temporal and cross-layer dynamics. This paper introduces a novel Tensor State Space Model for Dynamic Multilayer Networks (TSSDMN), utilizing a latent space model framework. TSSDMN employs a symmetric Tucker decomposition to represent latent node factors, their interaction patterns, and layer transitions. Then by fixing the latent factors and allowing the interaction patterns to evolve over time, TSSDMN uniquely captures both the temporal dynamics within layers and across different layers. 
The model identifiability conditions are discussed. 
By treating the interactions of latent factors as variables whose posterior distributions are approximated using a mean-field variational inference approach, a variational Expectation Maximization algorithm is developed for efficient model inference. Numerical simulations and case studies demonstrate the efficacy of TSSDMN for understanding dynamic multilayer networks.
\end{abstract}

\begin{IEEEkeywords}
Dynamic Network model, Tensor decomposition, State space model, Multilayer network, Bayesian posterior estimation, Variational inference.
% Networked time series data, few-shot learning, meta-learning, Bayesian Hierarichical Model.
\end{IEEEkeywords}

% \fi
\IEEEpeerreviewmaketitle

\section{Introduction}
\label{sec:introduction}
Network modeling is important for describing and analyzing complex systems across social 
 \cite{liu2024controlling}, biological \cite{li2024robust}, information \cite{benbya2020complexity}, and engineering sciences \cite{meng2022operational}. Traditionally, these systems are represented as ordinary graphs, where nodes correspond to entities and edges indicate connections. However, as systems grow more complex, particularly when multiple types of connections exist, single-layer network models often fall short in capturing their full structure. This limitation has led to the development of multilayer networks, which provide a more expressive framework by incorporating multiple types of relationships. In a multilayer network, each layer represents a distinct type of connection. For example, in a time-stamped social network, interactions such as phone calls, text messages, emails, and face-to-face meetings can be represented in separate layers.

A key challenge in multilayer network modeling lies in balancing information aggregation and differentiation. Aggregating all layers into a single network may obscure critical layer-specific details, whereas analyzing each layer independently fails to exploit shared patterns across layers. Therefore, tailored analytical tools are needed to capture both common structures and layer-specific variations \cite{loyal2023eigenmodel}. Among various modeling approaches, latent space models (LSMs) have gained prominence. First introduced by \cite{hoff2002latent}, LSMs assume that nodes have some latent factors that influence how they connect with each other. This framework was later extended by \cite{hoff2007modeling} to a more general decomposition, to describe how the unobserved latent factors affect edge information. Recently, \cite{gollini2016joint,d2019latent,zhang2020flexible} extend LSM to multilayer scenarios, by considering the latent factors are shared across layers to capture commonalities across layers.  Yet each layer can have layer-specific bias and factor interaction patterns, accommodating distinctions between layers. These LSM-type models enhance flexibility and explainability in modeling heterogeneous network structures. Refer to Section \ref{sec:relatedwork} for a comprehensive review. 

The problem becomes more complex when the networks are temporally evolving. In such a case, it is essential to examine both \emph{intra-layer dynamics}, which describes how connections evolve within each layer, and \emph{cross-layer dynamics}, which captures how different layers influence each other's connection dynamics. For example, in a multilayer social network, connections between individuals are time-dependent. Two people who recently communicated may have a lower probability of reconnecting in the near future. Additionally, different communication modes interact: if two individuals recently exchanged emails, they are more likely to continue using this mode in the short term.

Despite growing interest in dynamic multilayer network modeling, existing latent space models (LSMs) struggle to flexibly capture both intra-layer and cross-layer dynamics. In particular, \cite{durante2017bayesian} develops a Bayesian model framework by decomposing the edge probabilities as a function of shared and layer-specific node factors in a latent space. Then, it models the dynamics of the layer-specific factors in each layer separately via Gaussian processes. \cite{rodriguez2022multilayered} further extends it by adding dynamics of shared node factors via a Gaussian process as well. \cite{loyal2023eigenmodel} also proposes to model the dynamics of shared node factors via a random walk model.  However, these models do not explicitly capture cross-layer dependencies, limiting their ability to model inter-layer interactions.

Furthermore, real-world interaction networks often exhibit clustered structures and local connectivity patterns, especially when the network is large-scale. In the communication network example, people who frequently interact can be grouped into distinct social communities. While community detection has been extensively studied in static multilayer networks and dynamic single-layer networks, relatively few works have addressed this point in dynamic multilayer networks.

This paper introduces a Tensor State Space model for Dynamic Multilayer Networks (TSSDMN) within the LSM framework. TSSDMN represents the log-odds of edge connection probabilities in a multilayer network as a tensor and employs symmetric Tucker decomposition to capture three key components: latent node factors, interaction patterns between factors within each layer, and layer-transition patterns. To ensure model identifiability, we impose nonnegativity constraints on the latent factors. This regularization also enhances interpretability, as nonnegative factors correspond to weights on specific interaction patterns, which can be interpreted as a direct community detection for the dynamic multilayer network. 

Based on Tucker decomposition, TSSDMN further incorporates a tensor state space model to characterize the temporal evolution of edge connections. Unlike existing approaches such as \cite{durante2017bayesian} and \cite{loyal2023eigenmodel}, which model latent factors as time-varying, our approach fixes the latent factors and instead allows factor interaction patterns to evolve over time. Considering these interaction patterns across multiple layers form a tensor, we adopt a tensor autoregressive model to capture both intra-layer and cross-layer dependencies,  enabling a more flexible and structured representation of dynamic multilayer networks. The identificability properties of TSSDMN are carefully discussed, and a variational Expectation Maximization algorithm is developed for model parameter estimation. 
%Specifically, we treat latent factors as unobserved latent variables and approximate their posteriors using mean-field variational inference. The remaining model parameters are estimated via a block coordinate descent algorithm, maximizing the expected marginal log-likelihood under the posterior of the latent factors. In addition, we propose a hyperparameter selection algorithm to determine the optimal latent factor dimension.

The remainder of this paper is structured as follows. Section \ref{sec:relatedwork} provides a detailed review of existing network modeling methods. Section \ref{sec:methodology} introduces TSSDMN, and discusses its identifiability properties and interpretability in the context of community detection. Section \ref{sec:model estimation} describes the model estimation procedure and hyperparameter selection algorithms. Section \ref{sec:experiments} presents numerical studies on synthetic data, followed by Section \ref{sec:case}, which applies TSSDMN to two real-world case studies. Finally, Section \ref{sec:conclusion} summarizes key findings and conclusions.

\section{Related Work}
\label{sec:relatedwork}
In recent years, statistical methods for network data analysis have seen significant growth, with LSMs and stochastic block models (SBMs) emerging as two dominant approaches. 

LSMs map each node to a lower-dimensional latent space, which determines the underlying connection probabilities between nodes. LSM interprets these latent factors as a node’s unmeasured factors such that nodes that have similar factors in the latent space are more likely to have connections. This interpretation naturally explains the high levels of homophily and transitivity in real-world networks. In contrast, SBMs assume that nodes belong to discrete communities and connections depend only on community membership. This formulation is inherently suitable for community detection. However, its hard community assignments make SBMs less effective in capturing node-specific factor information. In the following sections, we review existing studies on dynamic network and multilayer network modeling using these two approaches.

\subsection{Dynamic Network Modeling}
\label{sec:dynamic_review}
Dynamic network models analyze evolving relationships between nodes and capture network structural changes over time. In the latent space model (LSM) framework, most approaches assume that latent factors evolve dynamically, influencing connection probabilities.

Dynamic network models focus on analyzing the evolving relationships between nodes and capturing network structural changes over time. In the LSM framework, most approaches consider modeling latent factors as a dynamic process. For example, \cite{sarkar2005dynamic} and \cite{sewell2015latent} assume the connection probabilities between nodes as interactions between their latent factors, which evolves according to a Gaussian random walk model. \cite{heaukulani2013dynamic} considers that each latent factor has different states, and each node's state propagates according to a hidden Markov model. \cite{durante2014nonparametric} assumes the latent factors of all the steps joint follow a Gaussian process. \cite{mazzarisi2020dynamic} further proposes an autoregressive model to describe the dynamics of latent factors. \cite{kim2023dynamic} also consider the interaction patterns of latent factors including both additive and multiplicative effects \cite{hoff2021additive}. However, all the existing dynamic LSMs assume the factor interaction patterns do not change over time and hence have limited flexibility to model network dynamics in reality. 

Dynamic SBMs capture network evolution by allowing for changes in community structure, edge weights, and node attributes over time. \cite{yang2011detecting} presents a probabilistic framework for analyzing dynamic communities by allowing each node's community to switch over time. In contrast, \cite{xu2014dynamic} assumes fixed community memberships while allowing connection probabilities between communities to evolve according to a linear dynamic model. Combining both aspects, \cite{matias2017statistical} allows both node memberships and community connection probabilities to change over time.
An alternative variant of the dynamic SBM, known as the dynamic mixed-membership SBM, allows each node to belong to multiple communities simultaneously \cite{xing2010state}. 
These models represent a node’s community affiliations as a membership vector.  \cite{ho2014analyzing} assumes the mixed-membership vector for each node has a multinomial distribution, and each of its parameters after logistic transform follows a random walk model. \cite{olivella2022dynamic} assumes the mixed-membership vector follows a Markov-dependent mixture.
%\cite{fan2014dynamic} even generalizes it to cases with potentially infinite communities, allowing the number of communities to change over time. 
%\cite{olivella2022dynamic} proposes a novel approach to mixed-membership SBM by positing that the mixed memberships adhere to a hidden Markov process.
% \cite{bhattacharjee2020change} focus on the single change point in a dynamic SBM, where the change can be community structure or community connection probabilities. 

\subsection{Multilayer Network Modeling}
\label{sub:multilayer_review}
Several pioneering studies have extended single-layer latent space models (LSMs) to multilayer networks. Most of these methods assume that all layers share a common set of latent factors, while allowing layer-specific interaction patterns to differentiate connection probabilities across layers \cite{d2019latent}. For example, \cite{gollini2016joint} models the connection log-odds as a function of the Euclidean distance between latent factors, with layer-specific intercepts. \cite{zhang2020flexible} builds on this by incorporating both layer-specific intercepts and layer-specific factor interaction patterns, offering greater modeling flexibility. \cite{salter2017latent} extends the framework by using a multivariate Bernoulli likelihood to model cross-layer dependencies, thus capturing associations between layers. For large-scale networks, \cite{arroyo2021inference} imposes a low-rank structure on the multilayer connection log-odds matrix, assuming that its column space is spanned by shared latent factors. Similar approaches are adopted in \cite{wang2019joint} and \cite{jones2020multilayer}. Unlike the above models that typically assume the latent factors are shared across all the layers, \cite{macdonald2022latent} allows each layer to have its own latent factors, to better reserve layer-specific details. 

A few recent studies have extended multilayer latent space models (LSMs) to dynamic settings by integrating ideas from dynamic LSMs (as reviewed in Section \ref{sec:dynamic_review}). \cite{loyal2023eigenmodel} models the temporal evolution of shared latent factors using a random walk process. \cite{durante2017bayesian} introduces both shared and layer-specific latent factors, modeling their dynamics via Gaussian processes. Focusing specifically on community structure, \cite{rodriguez2022multilayered} assumes that each community, rather than each node, is associated with shared and layer-specific latent factors that evolve independently over time through layer-wise Gaussian processes, while node-to-community memberships remain fixed. However, these approaches have two key limitations. First, like earlier dynamic LSMs, they assume that interaction patterns among latent factors are fixed over time, which limits their ability to model intra-layer dynamics. Second, they do not account for cross-layer dependencies, as layer-specific factor evolution is treated independently across layers, preventing them from capturing cross-layer interactions.

In multilayer stochastic block models (SBMs), many studies assume that community structures are consistent across layers, while allowing edge connection probabilities between communities to vary by layer. For example, \cite{han2015consistent} proposes a multilayer SBM where edge probabilities depend on node community memberships and a layer-specific intercept. In contrast, \cite{stanley2016clustering} allows community structures to differ across layers by introducing a clustering step that groups layers into distinct strata, each modeled by a separate SBM. \cite{lei2023bias} extends spectral clustering to the multilayer SBM setting by summing squared adjacency matrices across layers and applying a bias adjustment. This approach enables the detection of shared community structures, even when individual layers lack sufficient signal on their own. 

Extensions of multilayer SBMs to dynamic settings remain limited. \cite{lopez2022dynamic} is among the first to allow for layer-specific community structures, where node memberships evolve over time via a hidden Markov model. However, compared to LSMs, SBMs are less suited for modeling continuous-time dynamics, as they typically rely on discrete community assignments and abrupt transitions, making them less flexible in capturing the smooth temporal evolution often observed in real-world networks.

\section{Methodology}
\label{sec:methodology}
\subsection{Model Definition}
\label{sec:model definition}
\begin{figure*}[t]
    \centering
    \includegraphics[width=\linewidth]{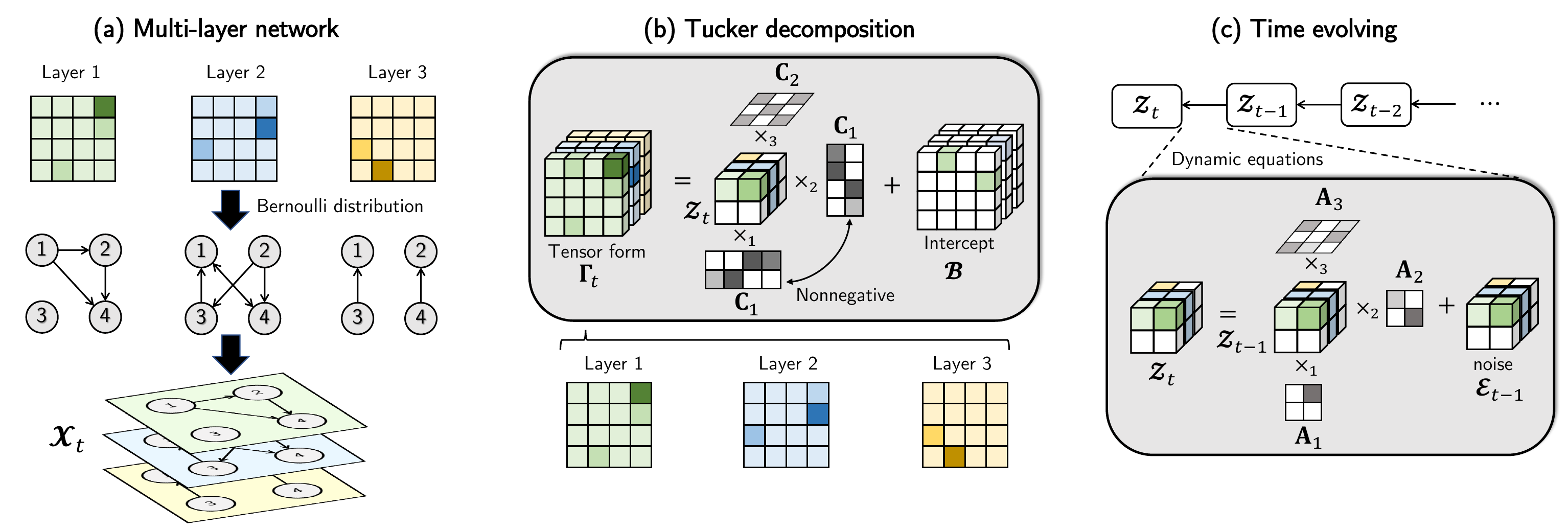}
    \vspace{-20pt}
    \caption{The overall framework of TSSDMN}\label{fig:methodology}
    \vspace{-10pt}
\end{figure*}

Let $n$ denote the number of nodes and $K$ the number of layers. At each time point $t \in \{1,2,\ldots,T\}$, we observe a dynamic multilayer network represented by a binary tensor $\bcalX_t\in \{0,1\}^{n \times n \times k}$, where $\calX_{t,ijk}=1$ indicates the presence of a directed edge from node $i$ to node $j$ in layer $k$ at time $t$. We assume each observed edge follows a Bernoulli distribution
\begin{equation} \label{equ:bern}
    \calX_{t,ijk} \sim {\rm Bernoulli}(p_{t,ijk}),
\end{equation}
where $p_{t,ijk} \in (0,1)$ denotes the edge probability.

To model these probabilities, we apply a logit transformation, which maps $p_{t,ijk}$ from the internal $(0,1)$ to the real line $\bbR$, yielding the log-odds $\gamma_{t,ijk}$:
\begin{equation}
\label{eq:latent_space}
    \gamma_{t,ijk} = \log\left(\frac{p_{t,ijk}}{1 - p_{t,ijk}}\right)=\sum_{k'=1}^{K}c_{2,k'k}\bc_{1,i}^{T}\bZ_{t,k'}\bc_{1,j}+b_{ijk},
\end{equation}
where $\bc_{1,i}=[c_{1,i1}, c_{1,i2}, \ldots, c_{1,il}]^T \in \bbR^m$ denotes the latent factor vector of node $i$, shared across all the layers. These factors capture intrinsic node characteristics that are consistent across layers. $\bZ_{t,k'}\in \bbR^{m \times m}$ is the factor interaction matrix for layer $k'$ at time $t$, where its $ll'$ component $Z_{t,k'll'}$ represents the interaction strength from factor $l$ to factor $l'$. $c_{2,k'k}$ is a cross-layer influence coefficient, quantifying how much the interaction structure in layer $k'$ contributes to the edge formation in layer $k$. 

Compared with existing LSMs for dynamic multilayer networks \cite{durante2017bayesian,loyal2023eigenmodel}, a key distinction of our approach lies in the modeling of dynamics. While previous methods typically assume that the latent node factors $\bc_{1}$ evolve over time, we instead assume that $\bc_{1}$ remains time-invariant, and the interaction patterns among latent factors, represented by $\bZ_{t,k'}$ are time-varying. This formulation offers two main advantages. First, it aligns more closely with real-world scenarios in which intrinsic node characteristics tend to be stable while interaction dynamics evolve over time. For example, in a social network, a latent factor may represent a preference such as ``email communication in the morning.'' Then, the interactions corresponding to this latent factor will be more intensive in the morning and weaker in other time, i.e., varying over time. Second, this modeling strategy provides better separation between the static and dynamic components of the network. 

Second,  our modeling strategy provides better separation between static and dynamic components of the network. By fixing $\bc_{1}$ and incorporating a static bias term $\calB$, the model allows $\bZ_{t,k}$ to more effectively capture the dynamic structure of the network. As a result, the latent factors $\bc_{1}$ can concentrate on encoding meaningful, temporally stable attributes, while the temporal evolution is driven by the changing interaction patterns.

An additional contribution of our model is the introduction of the cross-layer influence parameter $c_{2,k'k}$, which explicitly captures the effect of inter-layer dependencies—i.e., how the interaction structure in one layer influences connection behavior in another. This allows the model to flexibly account for cross-layer dynamics, which are common in multilayer relational systems.

Furthermore, we impose a nonnegativity constraint on the latent factors $\bc_{1,i}$, ensuring that each component to be $c_{1,il}\geq 0$. Given that the interaction term can be expressed as 
\begin{align}
\bc_{1,i}^{T}\bZ_{t,k'}\bc_{1,j}=\sum_{l=1}^{m}\sum_{l'=1}^{m}c_{1,il}Z_{t,k'll'}c_{1,il'}.
\end{align}
This regularization offers two key benefits. First, nonnegativity provides an intuitive interpretation of the latent factors: each element $c_{1,il}$ quantifies the strength of latent factor $l$ associated with node $i$, and the log-odds $\gamma_{t,ijk}$ becomes an additive composition of latent factor interactions, facilitating model interpretability. Second, the nonnegativity constraint contributes to model identifiability, helping to avoid issues related to sign ambiguity and redundant representations in the latent space. In addition, we relax the common homophily assumption by not restricting $\bZ_{t,k'}$ to be a diagonal matrix. Instead, we allow general interactions between different latent factors, which increases the expressiveness of the model and enables it to capture heterophily and complex cross-factor influences that may exist in real-world networks.

Let $\bGamma_t=[\gamma_{t,ijk}] \in \bbR^{n \times n \times K}$ denote the log-odds tensor that encodes edge probabilities across all layers at time $t$. The latent factor model in Eq. (\ref{eq:latent_space}) can be reformulated as a Tucker decomposition of the log-odds tensor: 
\begin{equation} \label{equ:beta}
    \bGamma_{t} = \bcalZ_{t} \times_1 \bC_1 \times_2 \bC_1 \times_3 \bC_2 + \bcalB. % \sum_{i=1}^m \sum_{j=1}^m \sum_{k=1}^K \bC_{1,ii'} \bC_{1,jj'} \bC_{2,kk'} \bcalZ_{t,i'j'k'} + \bcalB_{ijk},
\end{equation}
Here $\bC_1 = [\bc_{1,i}^T, \bc_{2,i}^T, \ldots, \bc_{n,i}^T]^T \in \bbR^{n \times m}$ is the matrix of latent factors of all the nodes. $\bC_2 = [c_{2,k'k}] \in \bbR^{K \times K}$ is the layer-transition matrix which specifies cross-layer interactions. $\bcalZ_t \in \bbR^{m \times m \times K}$  is the core tensor capturing time-varying interaction patterns among latent factors in each layer, and $\bcalB \in \bbR^{n \times n \times K}$ is a static bias tensor, accounting for residual structures not captured by the latent factor interactions. The Tucker formulation compactly expresses the multi-aspect interactions between nodes, factors, and layers. 

To capture the temporal dynamics of the network, we assume that the core tensor $\calZ_{t}$ evolves over time, according to a third-order tensor autoregressive model:
\begin{equation} \label{equ:tensor-regression}
 \calZ_{t,ijk} = \sum_{i'=1}^m \sum_{j'=1}^m \sum_{k'=1}^K A_{1,i'i} A_{2,j'j}A_{3,k'k} \calZ_{t-1,i'j'k'} + \varepsilon_{t,ijk},
\end{equation}
where $A_{1,i'i}$ models the temporal influence from latent factor $i'$ to $i$ in the outgoing mode.  $A_{2,j'j}$ models the temporal influence from latent factor $j'$ to factor $j$ in the incoming mode. $A_{3,k'k}$ captures the cross-layer temporal correlation from layer $k'$ to layer $k$. $\varepsilon_{t,ijk} \sim \calN(0, \sigma^2)$ represents Gaussian noise at time $t$. The initial state $\bcalZ_0$ is assumed to follow an independent Gaussian distribution $\calZ_{0,ijk} \sim \calN(\calU_{0,ijk}, \omega^2)$, where $\calU_{0,ijk}$ and $\omega^2$ denote the mean and variance parameters, respectively. 

Equation \eqref{equ:tensor-regression} can be compactly reformulated using the Tucker product as: 
\begin{equation}
\label{equ:Z_tensor}
    \bcalZ_{t} = \bcalZ_{t-1} \times_1 \bA_1 \times_2 \bA_2 \times_3 \bA_3 + \bcalE_t,
\end{equation}
where $\bA_1 =[A_{1,i'i}] \in \bbR^{m\times m}$, $\bA_2=[A_{2,j'j}] \in \bbR^{m\times m}$,
$\bA_3=[A_{3,k'k}] \in \bbR^{K \times K}$, 
and $\bcalE_t =[\varepsilon_{t,ijk}] \in \bbR^{m \times m \times K}$.

This formulation allows the model to simultaneously capture intra-layer dynamics within the latent space and cross-layer dynamics, offering a flexible and structured approach to modeling evolution in dynamic multilayer networks. The overall framework of TSSDMN is shown in Figure \ref{fig:methodology}. 

\subsection{Identifiability Property}
\label{sec:identifiable}
Traditional LSMs are known to suffer from inherent unidentifiability, as their representations are invariant under arbitrary linear transformations of the latent factors \cite{walter2013identifiability}. This ambiguity undermines the interpretability of the learned embeddings and complicates model analysis. Our nonnegative Tucker decomposition addresses this limitation through constrained parameterization. Specifically, we restrict the latent factor matrix to be nonnegative, which eliminates rotational ambiguity and facilitates meaningful factor interpretation.

We begin by formalizing the concept of observational equivalence, which characterizes when two different parameter sets yield identical distributions over observable data. 
%We first formalize the concept of observational equivalence in Definition \ref{def:observational}, then establish identifiability guarantees under reasonable assumptions.
\begin{definition}[Observational equivalence]
\label{def:observational}
Two parameter sets $\Theta=\{\bA_1, \bA_2, \bA_3,\bC_1, \bC_2, \bcalB, \bcalU_0, \sigma^2, \omega^2\}$ and $\Theta'=\{\bA'_1, \bA_2', \bA_3', \bC'_1, \bC_2', \bcalB',\bcalU'_0, \sigma^{2'}, \omega^{2'}\}$ are said to be observationally equivalent if for all $t=1,2,\ldots$ and for any observable tensor $\calX_{t} \in \{0,1\}^{m\times m \times K}$, the corresponding likelihoods are identical:
    \begin{equation*}
        p(\calX_{t};\Theta) \equiv p(\calX_t;\Theta').
    \end{equation*}
\end{definition}
To establish identifiability, we impose a mild stationarity condition on the tensor autoregressive process, ensuring asymptotic stability of the dynamic components.

\begin{assumption}[Stationarity] 
\label{ass:stationarity}
The tensor autoregressive process satisfies the following spectral radius conditions: 
\begin{align}
\rho(\bA_i) < 1, \text{ and } \rho(\bA_i') < 1 \text{ for all } i=1,2,3, 
\end{align}
where $\rho(\cdot)$ denotes the sepctral radius of a matrix.These conditions ensure that the process is asymptotically stable over time.
\end{assumption}

Assumption \ref{ass:stationarity} ensures convergence of the system dynamics, thereby enabling the unique recovery of the static network parameters from the observed data. 
%Our identifiability analysis proceeds in two stages. Theorem \ref{the:static} establishes the identifiability of the static bias $\bcalB$ under the stability condition of Assumption \ref{ass:stationarity}. Building on this result, Theorem \ref{the:identifiability} addresses the identifiability of the dynamic components, given additional structural constraints on the latent space.

% \begin{theorem}[Identifiability of Static Bias]
% \label{the:static}
% Let $\Theta$ and $\Theta'$ be two observationally equivalent parameter sets that satisfy Assumption \ref{ass:stationarity}. Then their corresponding static bias tensors are equal $\bcalB = \bcalB'$.
% \end{theorem}
% \proof
% See Appendix \ref{appx:proof}.
% \endproof

To further establish full model identifiability including the latent factors, we introduce a common structural constraint from the nonnegative matrix and tensor factorization literature:

%The stability condition in Assumption \ref{ass:stationarity} guarantees the convergence of system dynamics, enabling unique recovery of static network parameters. To achieve full model identifiability, we introduce a structural constraint common in nonnegative tensor factorization:

\begin{assumption}[Pure-source Dominance]
\label{ass:pure-source}
The latent membership matrix $\bC_1$ admits a permuted anchored structure:
\begin{equation*}
    \bC_1 = \bPi_1\begin{bmatrix}
        \bI_m \\
        \mathbf{U}
    \end{bmatrix}\bPi_2 \bD_0,
\end{equation*}
where $\bPi_1 \in \bbR^{n \times n}$ and $\bPi_2 \in \bbR^{m \times m}$ are permutation matrices, $\bD_0 \in \bbR^{m \times m}$ is a diagonal scaling matrix, and $\mathbf{U} \in \bbR^{(n-m) \times m}$ is an arbitrary matrix. This structure ensures that each latent dimension is anchored by at least one ``pure'' node, that is, a node that is exclusively associated with a single latent factor.
\end{assumption}

Assumption \ref{ass:pure-source} is widely used in nonnegative matrix and tensor decomposition for ensuring identifiability \cite{robinson2009non, zhou2015efficient}. In the context of networks, it corresponds to requiring that each latent behavioral pattern has at least one prototypical node that expresses this pattern exclusively, thereby grounding the latent semantics.

\begin{theorem}[Model Identifiability] \label{the:identifiability}
Let $\Theta$ and $\Theta'$ be two observationally equivalent parameter sets that satisfy Assumptions \ref{ass:stationarity} and \ref{ass:pure-source}, and further assume that 
\begin{align}
\|\bC_1\|_F = \|\bC'_1\|_F, \|\bC_2\|_F = \|\bC'_2\|_F. 
\end{align}
Then the following hold: 
\begin{enumerate}
    \item $\bcalB=\bcalB'$.
    \item There exists a permutation matrix $\bPi \in \bbR^{m \times m}$ such that $\bC_1 = \bC'_1 \bPi$.
    \item There exists an orthogonal matrix $\bR \in \bbR^{m \times m}$ such that $\bC_2 = \bC'_2 \bR$.
\end{enumerate}
Moreover, the dynamic interaction tensor satisfies the equivalence:
\begin{equation*}
    \bcalZ_t = \bcalZ'_t \times_1 \bPi \times_2 \bPi \times_3 \bR, \quad \forall t=1,\ldots,T.
\end{equation*}
\end{theorem}
\proof
See Appendix \ref{appx:proof-identifiability}.
\endproof

Theorem \ref{the:identifiability} guarantees the identifiability of the latent factor matrix $\bC_1$ up to a column permutation. The permutation matrix $\bPi$ simple reorders the latent dimensions of $\bC_1$, preserving their semantic consistency across models.  Therefore, each column of $\bC_1$ can be interpreted as a distinct and consistent latent factor. Furthermore, Theorem \ref{the:identifiability} shows that the layer-transition matrix $\bC_{2}$ is identifiable up to an orthogonal transformation. Consequently, the dynamic core tensor $\bZ_{t,k}$ is recoverable up within-layer permutation and cross-layer orthogonal transformation, ensuring the meaningful recovery of both temporal and cross-layer dynamics.

\begin{remark}[Community Detection based on $\bC_{1}$]
\label{rem:clustering}  
The identifiability results established in Theorem \ref{the:identifiability} enable interpretable community detection for the dynamic components of the network. In particular, the matrix of latent node factors $\bC_1$ which is identifiable up to column permutation, provides a consistent and interpretable representation of nodes in a shared latent space across time. Each column of $\bC_1$ corresponds to a latent community, and its temporal evolution is governed by the time-varying interaction patterns encoded in $\bcalZ_t$. 

The use of nonnegative matrix factorization (NMF) for graph community detection has been extensively studied, particularly in the context of weighted graphs \cite{kuang2012symmetric}. In this setting, the adjacency matrix $\bW$ is approximated via symmetric nonnegative matrix factorization, i.e., $\bW \approx \bC \bC^{T}$ where $\bC$ encodes the community membership strengths for each node. Building on this idea, Symmetric Nonnegative Matrix Tri-Factorization (SNMTF) \cite{wang2011community} generalizes the decomposition by introducing a symmetric community interaction matrix $\bZ$ such that $\bW = \bC \bZ \bC^{T}$. This formulation allows for modeling not only the membership of nodes but also the inter-community connectivity structure. More recently, \cite{zhang2023multi} extends SNMTF to multilayer networks by factorizing each layer's adjacency matrix as $\bW^{k} = \bC \bZ^{k} \bC^{T}$ and applying regularization to encourage similarity of interaction patterns $\bZ^{k}$ across layers. From this perspective, our proposed TSSDMN model (Eq. \ref{equ:beta}) can be viewed as a tensor-based generalization of SNMTF, where the shared node factor matrix $\bC_{1}$ plays the role of $\bC$, and the dynamic core tensor $\bcalZ_{t}$ captures time-varying, layer-specific interaction patterns. The matrix $\bC_{2}$ further introduces a learned mechanism for cross-layer interaction, automatically adapting the influence of each layer, thus enhancing the flexibility and expressiveness of the model.
\end{remark}
% \chen{maybe a better way is to add a new section 4: Temporal Dynamics Estimation: 4.1 VB for estimation; 4.2 error bound; change name of Section 5 to ``Model Inference via EM''}
\subsection{Convergence Property of $\bGamma_t$}
Under the identifiability condition of $\Theta$, in this section, we further establish the estimation error bounds for the dynamic core tensor $\{\bGamma_t\}_{t=1}^T$ in the proposed TSSDMN model. Our analysis comprises two main parts. First, we derive a minimax lower bound to quantify the fundamental statistical difficulty of the estimation problem. This bound serves as a benchmark for the optimal performance achievable by any estimator. Second, we establish an upper bound on the convergence rate of a Bayesian posterior estimator of $\bGamma_t$ determined by $\bcalZ_t$. These two bounds allow us to formally assess the statistical optimality of our approach.

To establish the minimax lower bound, a rigorous characterization of the parameter space is required. To enforce structured evolution and prevent overfitting to temporal noise, we regularize the dynamic component. This is formalized by constraining the total variation of the core tensor sequence, as specified in Definition \ref{def:tds}.

\begin{definition}[Dynamic Smoothness Constraint]
\label{def:tds}
The parameter space for the core tensor sequence is defined as:
\resizebox{1.03\columnwidth}{!}{
$\begin{aligned}
\text{TDS}(L) = \left\{ \{\boldsymbol{\mathcal{Z}}_t\}_{t=1}^T: \sum_{t=2}^T \|\boldsymbol{\mathcal{Z}}_t - \boldsymbol{\mathcal{Z}}_{t-1} \times_1 \mathbf{A}_1 \times_2 \mathbf{A}_2 \times_3 \mathbf{A}_3\|_F^2 \le L \right\}
\end{aligned}$,
}
where $L$ is a budget parameter controlling the degree of temporal smoothness.
\end{definition}

Furthermore, to ensure the stable and unique recovery of the latent structure, the factor matrices must be non-degenerate, preventing the latent space from collapsing. This is formalized in Assumption \ref{ass:non-degenerate-factors-summary}.

\begin{assumption}[Non-degenerate Factors]
\label{ass:non-degenerate-factors-summary}
The factor matrices $\mathbf{C}_1 \in \mathbb{R}^{n \times m}$ and $\mathbf{C}_2 \in \mathbb{R}^{K \times K}$ have bounded singular values. That is, there exist constants $0 < \lambda_{1\min} \le \lambda_{1\max} < \infty$ and $0 < \lambda_{2\min} \le \lambda_{2\max} < \infty$ such that:
\begin{gather*}
\lambda_{1\min} \le \sigma_{\min}(\mathbf{C}_1) \le \sigma_{\max}(\mathbf{C}_1) \le \lambda_{1\max}, \\
\lambda_{2\min} \le \sigma_{\min}(\mathbf{C}_2) \le \sigma_{\max}(\mathbf{C}_2) \le \lambda_{2\max}, \\
\lambda_{1,\min}\asymp \lambda_{1,\max}\asymp \sqrt{n}\text{ and }\lambda_{2,\min}\asymp \lambda_{2,\max}\asymp \sqrt{K}.
\end{gather*}
\end{assumption}

With the parameter space thus defined, we present the minimax lower bound for estimating the core tensor sequence.

\begin{theorem}[Minimax Lower Bound]
\label{the:minimax-lower-bound-summary}
Suppose the data is generated according to the TSSDMN model with parameters satisfying Assumptions \ref{ass:stationarity} to \ref{ass:non-degenerate-factors-summary}, then the minimax risk for estimating the core tensor sequence is lower-bounded as follows:
\begin{equation*}
\begin{split}
& \inf_{\{\hat{\boldsymbol{\mathcal{Z}}}_t\}} \sup_{\{\boldsymbol{\mathcal{Z}}_t\} \in \text{TDS}(L)} \mathbb{E} \left[ \frac{1}{Tn^2K} \sum_{t=1}^T \|\mathbf{\Gamma}_t - \hat{\mathbf{\Gamma}}_t\|_F^2 \right] \\
& \qquad \gtrsim \min \left\{\frac{m^2}{n^2T}, \frac{L^{2/3}m^{2/3}}{T^{2/3}n^{4/3}K^{1/3}}\right\},
\end{split}
\end{equation*}
where $\hat \bGamma_t$ represents an arbitrary estimator of $\bGamma_t$ determined by its corresponding $\hat \calZ_t$.
% \chen{where $\hat{\mathbf{\Gamma}}_t$ is ...}
\end{theorem}
\proof
See Appendix \ref{appx:minimax}
\endproof

Theorem \ref{the:minimax-lower-bound-summary} establishes a fundamental barrier on the estimation precision for the TSSDMN model, a result that holds for any conceivable estimator. 

% The bound is determined by the minimum of two terms. The first term, $\frac{m^2}{n^2T}$, dominates in the high-smoothness regime where the temporal variation budget $L$ is small. In this scenario, the problem behaves akin to estimating a nearly-static tensor. The error rate intuitively scales quadratically with the latent dimension $m$, reflecting the growing model complexity, and decreases with the number of nodes $n$ and time points $T$ as more information becomes available. The second term, $\frac{L^{2/3}m^{2/3}}{T^{2/3}n^{4/3}K^{1/3}}$, characterizes the error in the intermediate regime where temporal dynamics are prominent. The dependence on $L^{2/3}$ is characteristic of non-parametric problems, where estimating a smooth function (the tensor's trajectory) is inherently more challenging than estimating a fixed parameter. Notably, the error decreases faster with network size ($n^{-4/3}$) but slower with time ($T^{-2/3}$) compared to the static case, highlighting the unique challenges of dynamic estimation. The weak dependence on the number of layers ($K^{-1/3}$) suggests that while information is pooled across layers to refine the estimate, the benefit exhibits diminishing returns.

Having established a minimax lower bound in Theorem \ref{the:minimax-lower-bound-summary}, which quantifies the fundamental statistical difficulty of the problem for any estimator, we now turn to analyzing the performance of a proposed Bayesian estimation procedure for $\bcalZ_t$ and the corresponding $\bGamma_t$ . We show that its convergence rate, i.e., an upper bound on its error, matches this fundamental lower bound.

In the Bayesian paradigm, the quality of an estimator is assessed through its posterior contraction rate. This rate quantifies how quickly the posterior distribution concentrates its mass around the true data-generating parameters as the amount of data increases. The following theorem formally establishes this posterior contraction rate for our TSSDMN model.
%providing the key component for assessing the optimality of our approach,$P\left( \frac{1}{n^2KT} \sum_{t=1}^{T} d'(p(\mathcal{X}_{1:T}|\hat\bcalZ), p(\mathcal{X}_{1:T}|\bcalZ^*)) \leq M \epsilon^2_{n,m,K,T} \right) \rightarrow 1$
\begin{theorem}[Posterior Convergence Rate]
\label{the:convergence}
Suppose the data-generating process has the true core tensor values $\calZ_{t}^{*}$, which satisfies Assumption \ref{ass:non-degenerate-factors-summary} and the following assumptions for some constants $C, C_0 > 0$:
    \begin{enumerate}
        \item[(1)] $||\calZ_{t}^{*}||_F\leq C$, for all $t=1,\ldots,T$.
        \item[(2)] $||\bcalZ_{t}^{*}-\bcalZ_{t-1}^{*} \times \bA_1\times \bA_2 \times \bA_3||_{F}\leq \frac{C_0L}{T}$, for all $t=2,\ldots, T$, with $L=o(m^2KT)$.
    \end{enumerate}
Suppose $K$ is a known constant. Let $\epsilon_{n,m,K,T}=L^{\frac{1}{3}}m^{\frac{1}{3}}T^{-\frac{1}{3}}n^{-\frac{2}{3}}K^{-\frac{1}{6}}+\sqrt{\frac{m^2 \log{((n^2T)/m^2)}}{n^2T}}$. As $n^2KT \rightarrow \infty$, we have:
\begin{equation}
P\left(\frac{1}{n^2KT} d'(P_{\hat\calZ}, P_{\calZ^*})\leq M \epsilon^2_{n,m,K,T}\right) \rightarrow 1,
\end{equation}
where $M$ is a sufficiently large constant, $\hat\bcalZ=\{\hat\bcalZ_t\}_{t=1}^T$ and $\bcalZ^*=\{\bcalZ_t^*\}_{t=1}^T$. Here, $P_{\hat\calZ}$ denotes the data-generating distribution of $\mathcal{X}_{1:T}$ under parameter $\hat\bcalZ$ with density $p(\cdot \mid \hat\bcalZ)$ with $\hat\calZ_t$ sampled from posterior distribution, and $P_{\calZ^*}$ denotes the true data-generating distribution with density $p(\cdot \mid \bcalZ^*)$. The Hellinger distance $d'$ is taken between these data-generating distributions. Specifically,
$
d'^2(P_{\hat\bcalZ}, P_{\bcalZ^*}) = \int \Big(\sqrt{p(\mathcal{X}_{1:T}|\hat\bcalZ)} - \sqrt{p(\mathcal{X}_{1:T}|\bcalZ^*)}\Big)^2 \, {\rm d} \mathcal{X}_{1:T}.$ 
\end{theorem}
\proof
See Appendix \ref{appx:proof-convergence}.
\endproof

% Theorem \ref{the:convergence} provides a crucial theoretical guarantee for our  inference procedure, establishing that the posterior distribution contracts around the true data-generating parameters $\bcalZ^*$. This result ensures the statistical consistency of our method. The convergence rate, $\epsilon_{n,m,K,T}^2$, decomposes into two terms that mirror the distinct statistical challenges of the model. The first term, $L^{\frac{2}{3}}m^{\frac{2}{3}}T^{-\frac{2}{3}}n^{-\frac{4}{3}}K^{-\frac{1}{3}}$, quantifies the error from estimating the temporal dynamics. This rate insightfully captures the trade-off: the error increases with the temporal variation budget $L$ (allowing more complex dynamics) but decreases as more data is gathered over time $T$, across nodes $n$, and through layers $K$. The second term, $\frac{m^2 \log(n^2T/m^2)}{n^2T}$, reflects the error from estimating the static model components, akin to the challenge in standard latent space models. The overall rate is governed by the slower of these two, showcasing the estimator's ability to adapt to the dominant source of statistical complexity.

Establishing the optimality of our estimator requires connecting the posterior contraction rate (Theorem \ref{the:convergence}) with the minimax risk (Theorem \ref{the:minimax-lower-bound-summary}). In particular, on the one hand, Theorem \ref{the:minimax-lower-bound-summary} indicates the fundamental lower bound of any estimator. On the other hand, since the Hellinger distance $d'(P_{\hat\bcalZ},P_{\bcalZ^*})$ is asymptotically equivalent to the squared Frobenius norm of $\bGamma_t$, i.e., $d'(P_{\hat\calZ}, P_{\calZ^*})^2 \asymp \|\boldsymbol{\hat\Gamma_t} - \boldsymbol{\Gamma_t}^*\|_F^2$, Theorem \ref{the:convergence} reveals a powerful alignment between the Bayesian posterior estimator's upper bound and the fundamental lower bound, confirming that no other estimator can achieve a fundamentally faster rate of convergence for this problem class. Thus, it provides a strong theoretical endorsement of our model's design and the effectiveness of using Bayesian posterior estimator to robustly capture complex latent dynamics from tensor time series data. This guides our following model estimation method.

\section{Model estimation}
\label{sec:model estimation}
This section details the inference procedure for TSSDMN. First, we can write it using an equivalent vector formulation. To facilitate parameter estimation and computation, we first reformulate the model in an equivalent vectorized form. Let $\bcalX_t$ denote the observed multilayer network at time $t$. We vectorized it into a vector $\bx_t \in \bbR^{n^2K}$ by stacking its elements such that $\bx_{t,(i-1)nK + (j-1)K + k} = \bcalX_{t,ijk}$. The same vectorization is applied to the log-odds tensor $\bGamma_t$, static bias tensor $\bcalB$, dynamic cor tensor $\bcalZ_t$, and the initial core tensor mean $\bcalU_0$. Consequently, we obtain the vectors $\bgamma_t \in \bbR^{n^2K}$ with $\bgamma_{t,(i-1)nK + (j-1)K + k} = \bGamma_{t,ijk}$, $\bb \in \bbR^{n^2K}$ with $\bb_{t,(i-1)nK + (j-1)K + k} = \bcalB_{t,ijk}$, $\bz_t  \in \bbR^{m^2K}$ with $\bz_{t,(i-1)mK + (j-1)K + k} = \bcalZ_{t, ijk}$, and $\bu_0 \in \bbR^{m^2K}$ with $ \bu_{0,(i-1)mK + (j-1)K + k} = \bcalU_{0, ijk}$. 

Using these vectorized representations, we can reformulate Eq. (\ref{equ:beta}) and Eq. (\ref{equ:Z_tensor}) into vector forms as:
\begin{equation}
\begin{split}
    \bgamma_t &= \bC \bz_t + \bb, \\
    \bz_t &= \bA \bz_{t-1} + \boldsymbol{\varepsilon}_t, 
\end{split}
\end{equation}
where $\bC = \bC_2 \otimes \bC_1 \otimes \bC_1 \in \bbR^{n^2K \times m^2K}$, and $\bA = \bA_3 \otimes \bA_2 \otimes \bA_1 \in \bbR^{m^2K \times m^2K}$. "$\otimes$" refers to Kronecker product. $\ \boldsymbol{\varepsilon}_t \sim \calN(\mathbf{0}, \sigma^2 \bI_{m^2K})$. $\bz_0 = \bu_0 + \boldsymbol{\varepsilon}_0$ with $\ \boldsymbol{\varepsilon}_0 \sim \calN(\mathbf{0}, \omega^2 \bI_{m^2K})$.

Denote the full parameter set as $\Theta =  \{ \bC_1, \bC_2, \bA_1, \bA_2, \bA_3, \bb, \bu_0, \sigma^2, \omega^2 \}$. We treat the latent dynamic interactions $\bZ=[\bz_1, \ldots, \bz_T]\in \bbR^{m^2K \times T}$ as unobserved variables and use an Expectation-Maximization estimation framework to jointly estimate $\bZ$ and $\Theta$. 

The EM algorithm iteratively updates $\Theta^{(v+1)}$ in the $v$th iteration by treating $\bZ$ as ``missing data'' and maximizing the expected complete-data log-likelihood. The key insight is that maximizing the complete-data likelihood $p(\bX, \bZ|\Theta)$ where $\bX=[\bx_1, \bx_2, \ldots, \bx_T]$. The key insight is that maximizing the complete-data likelihood is more tractable than maximizing the marginal likelihood $p(\bX |\Theta)$ 

Specificially, given the Markovian structure of TSSDMN, the complete data log-likelihood has the form:
\begin{equation}
\label{eq:full_likelihood}
    \begin{split}
         &\log p(\bX, \bZ|\Theta) \\
         & = \log p(\bz_0|\Theta)+\sum_{t=1}^T \log p(\bx_t|\bz_t, \Theta) + \sum_{t=1}^{T}\log p(\bz_t|\bz_{t-1},\Theta) \\
      &= - \frac{1}{2\omega^2}(\bz_0 - \bu_0)^T(\bz_0 - \bu_0) \\
      & \quad - \frac{1}{2\sigma^2} \sum_{t=1}^T (\bz_t - \bA \bz_{t-1})^T(\bz_t - \bA \bz_{t-1}) \\
    & \quad + \sum_{t=1}^T \sum_{j=1}^{n^2K} \log \left(\frac{\exp \bx_{t,j}\bgamma_{t,j}}{1 + \exp \bgamma_{t,j}}\right) - \frac{1}{2} m^2K \log \omega^2 \\ 
    &\quad - \frac{1}{2} Tm^2K \log \sigma^2 + constant. 
\end{split}
\end{equation}
Since $\bZ$ is unavailable, EM instead maximize the expected complete-data log-likelihood, conditioned on the observed data and current parameter estimates $\Theta^{(v)}$. This defines the E-step of the EM algorithm:  
\begin{equation}
\label{eq:E-step}
\begin{split}
        \textbf{E Step: } \quad Q_n(\Theta, \Theta^{(v)}) &= \bbE_{\bZ|\bX,\Theta^{(v)}} \log p(\bX,\bZ|\Theta) \\
    &= \int \log p(\bX,\bZ|\Theta) p(\bZ|\bX, \Theta^{(v)}) {\rm d}\bZ 
\end{split}
\end{equation}
where $p(\bZ|\bX,\Theta^{(v)})$ is the posterior estimator of $\bZ$ given current parameter $\Theta^{(v)}$.

In the M-step, we seek the next parameter estimate $\Theta^{(v+1)}$ by solving the following constrained optimization problem:
\begin{equation}
\label{eq:M-step}
\begin{split}
\textbf{M Step: } \quad  \Theta^{(v+1)} &= \underset{\Theta}{\arg \max}\  Q_n(\Theta, \Theta^{(v)})\quad \quad \quad \quad \quad  \\
s.t. & \quad  \bC_{1,ij} \geq 0, \forall i,j.
\end{split}
\end{equation}
We talk about these two steps in detail as follows. 
\subsection{E-step} 
\label{sub:Estep}
Consider the posterior distribution $p(\bZ|\bX,\Theta^{(v)})$ of Eq. (\ref{eq:E-step}) is intractable, i.e., does not have a closed form. We adopt a mean-field variational inference algorithm to approximate it via a simpler tractable distribution $q(\bZ)$. The mean-field approximation assumes a fully factorized form for the variational distribution, i.e., $q(\bZ) = \prod_{t=0}^{T} q_t(\bz_t)$ where each $q_t(\bz_t)$ approximates the marginal posterior of the latent state at time $t$. 
The optimal $q(\bZ)$ is obtained by minimizing 
the Kullback-Leibler divergence between $q(\bZ)$ and $p(\bZ|\bX,\Theta^{(v)})$. 
Compared to sampling-based methods such as Markov Chain Monte Carlo (MCMC), variational inference offers a compelling trade-off between computational efficiency and accuracy. It often achieves comparable estimation quality while being significantly faster, particularly for high-dimensional latent variable models \cite{wang2018frequentist}.

We assume $q_t(\bz_t)$ follows a multivariate Gaussian distribution with mean $\tbmu_t \in \bbR^{m^2K}$ and a diagonal covariance matrix $\tsigma^2_t \bI \in \bbR^{m^2K \times m^2K}$, i.e.,
\begin{equation*}
    q_t(\bz_t) = \prod_{j=1}^{m^2K} \frac{1}{\sqrt{2\pi \tsigma^2_t}}\exp \left(-\frac{1}{2\tsigma^2_t} (\bz_{t,j} - \tbmu_{t,j})^2 \right),
\end{equation*}
Here, $\{\tbmu_t\}_{t=0}^T$ and $\{\tsigma^2_t\}_{t=0}^T$ are the variational parameters to be optimized.

Minimizing Kullback-Leibler divergence between $q(\bZ)$ and $p(\bZ|\bX,\Theta^{(v)})$ is equivalent to maximizing the evidence lower bound (ELBO), which can be represented as:
\begin{equation}
    {\rm ELBO} = \mathbb{E}_q(\log p(\bZ|\Theta^{(v)})) + \mathbb{E}_q(\log p(\bX|\bZ,\Theta^{(v)}))  - \mathbb{E}_q(\log q(\bZ)).
\end{equation}
In particular, 
\begin{align*}
    &\mathbb{E}_q{\log p(\bZ|\Theta^{(v)})} \\
    &= - \frac{1}{2\omega^{2 (v)}} (\tbmu_0 - \bu_0^{(v)})^T(\tbmu_0 - \bu_0^{(v)}) - \frac{1}{2\sigma^{2 (v)}} \sum_{t=1}^T \| \tbmu_{t} - \bA^{(v)} \tbmu_{t-1} \|_2 ^2 \\
    & \quad - m^2K\frac{\tsigma_0^2}{2 \omega^{2(v)}} - \frac{1}{2\sigma^{2 (v)}}  \sum_{t=1}^T {\rm tr}(\tsigma^2_t \bI + \bA^{(v)} (\bA^{(v)})^T \tsigma_{t-1}^2) \\
    & \quad - \frac{1}{2} m^2K \log \omega^{2 (v)} - \frac{1}{2} Tm^2K \log \sigma^{2 (v)} + constant, 
\end{align*}
and
\begin{align*}
    \mathbb{E}_q (\log q(\bZ)) = - \frac{1}{2} \sum_{t=0}^T m^2K \log \tsigma^2_t + constant.
\end{align*}
For $\mathbb{E}_q (\log p(\bX|\bZ, \Theta^{(v)}))$, since it is difficult to compute the expectation of a logarithm function, we instead maximize its lower bound by Jensen's Inequality, denoted as $\underline{\mathbb{E}_q (\log p(\bX|\bZ, \Theta^{(v)}))}$:
\begin{align}
&\mathbb{E}_q (\log p(\bX|\bZ, \Theta^{(v)}))  \nonumber \\
    &= \mathbb{E}_q (\log(\prod_{t=1}^{T} \prod_{j=1}^{n^2K} \frac{\exp \bx_{t,j} \bgamma_{t,j}}{1 + \exp \bgamma_{t,j}}))  \nonumber \\
    % = \mathbb{E}_q (\sum_{t=1}^{T} \sum_{j=1}^{n^2K} \log(\frac{\exp \bx_{t,j} \bgamma_{t,j}}{1 + \exp \bgamma_{t,j}})) \\
    &\geq \sum_{t=1}^{T} \sum_{j=1}^{n^2K} \bx_{t,j}(\bC^{(v)} \tbmu_t + \bb^{(v)})_j - \log(1 + \mathbb{E}_q(\exp \bgamma_{t,j}))  \nonumber \\
    &= \sum_{t=1}^T \bx_t^T (\bC^{(v)} \tbmu_t + \bb^{(v)})  \nonumber \\
    &\quad- \sum_{t=1}^T \sum_{j=1} ^{n^2K} \log\left(1 + \exp \left((\bC^{(v)} \tbmu_t + \bb^{(v)})_j + \frac{\Tilde{\sigma}^2_t \bC^{(v)}_j\bC_j^{(v)T} }{2} \right) \right) \nonumber \\
    &:= \underline{\mathbb{E}_q (\log p(\bX|\bZ, \Theta^{(v)}))},
\label{eq:bound}
\end{align}
where $\bC_j^{(v)}$ is the $j$-th row of matrix $\bC^{(v)}$. 

Consequently, the lower bound of the ELBO is given by \(\underline{\text{ELBO}} = \underline{\mathbb{E}_q \log p(\mathbf{X}|\mathbf{Z}, \Theta^{(v)})} + \mathbb{E}_q\log p(\mathbf{Z}|\Theta^{(v)}) - \mathbb{E}_q\log q(\mathbf{Z})\). We instead maximize its lower bound \(\underline{\text{ELBO}}\) in the E-step, as this is computationally more feasible. This approach is theoretically justified since optimizing \(\underline{\text{ELBO}}\) still drives \( q(\mathbf{Z})\) towards minimizing the Kullback-Leibler divergence between it and the true posterior \( p(\mathbf{Z}|\mathbf{X}, \Theta^{(v)})\) ~\cite{drugowitsch2013variational}. Such maximization can be solved by blocked coordinate descent algorithm for each time step $t$. The detailed algorithm is shown in Appendix \ref{appx:m-step-details}.

\subsection{M Step}
In the M-step, our goal is to maximize 
\( Q_n^V(\Theta, \Theta^{(v)}) = \mathbb{E}_q \log p(\mathbf{X}, \mathbf{Z} | \Theta) = \mathbb{E}_q \log p(\mathbf{X}|\mathbf{Z}, \Theta) + \mathbb{E}_q \log p(\mathbf{Z} | \Theta) \). However, computing \(\mathbb{E}_q \log p(\mathbf{X}|\mathbf{Z}, \Theta) \) directly is challenging as mentioned in the E-step. To address this, we introduce a lower bound, \(\underline{\mathbb{E}_q \log p(\mathbf{X}|\mathbf{Z}, \Theta)}\) following the same derivation as Eq. (\ref{eq:bound}), and yield \(\underline{Q_n^V(\Theta, \Theta^{(v)})} = \underline{\mathbb{E}_q \log p(\mathbf{X}|\mathbf{Z}, \Theta)} + \mathbb{E}_q\log p(\mathbf{Z}|\Theta)\), which is more tractable to optimize. The M-step optimization problem is formulated as:
\begin{equation} 
\label{equ:objective-Mstep}
\begin{split}
    \Theta^{(v+1)} &= \underset{\Theta}{\arg \max} \ \underline{Q_n^V(\Theta, \Theta^{(v)})} \\
    & s.t. \quad \bC_{1,ij} \geq 0, \forall i,j.
\end{split}
\end{equation}
The parameter set $\Theta$ can be partitioned into two functionally independent subsets. The first group consists of $\bC_1$, $\bC_2$, and $\bb$, which determine the probabilistic link from $\bZ$ to $\bX$. The second group consists of $\bA_1$, $\bA_2$, $\bA_3$, $\bu_0$, $\omega^2$, and $\sigma^2$, which determine the dynamics of $\bZ$. Due to the separability of these two components in the objective function, we can optimize each subset independently, which significantly simplifies the M-step and improves computational efficiency.

\textbf{For the first group $\{\bC_1, \bC_2, \bb\}$}: to estimate $\bC_1$, we use projected gradient descent to enforce non-negativity:
\begin{equation}
\bC_{1,ij} \leftarrow \max \left\{0, \bC_{1,ij} + \alpha\left(\frac{\partial \underline{Q_n^V}}{\partial \bC_{1,ij}} \right) \right\},
\end{equation}
where $\alpha$ is the step size. To estimate $\bC_2$ and $\bb$, we can use gradient descent algorithms. 

\textbf{For the second group $\{\bA_1, \bA_2, \bA_3, \bu_0,\omega^2,\sigma^2\}$:} to estimate $\{\bA_1, \bA_2, \bA_3\}$, we can use gradient descent algorithms. $\bu_0$, $\omega^2$, and $\sigma^2$ have closed-form solutions:
\begin{align}
   \bu^{(v+1)}_0 &= \tbmu_0 ,\label{equ:update-u} \\ 
   \omega^{2(v+1)} &= \tsigma_0^2, \label{equ:update-omega} 
\end{align}
\begin{equation}
\begin{split}
   \sigma^{2(v+1)} &= \frac{1}{Tm^2K} \sum_{t=1}^T \| \tbmu_{t} - \bA^{(v+1)} \tbmu_{t-1} \|_2 ^2 \\
   & \quad + \frac{1}{Tm^2K} \sum_{t=1}^T {\rm tr}(\tsigma^2_t \bI + \bA^{(v+1)} \bA^{(v+1)T} \tsigma_{t-1}^2) . \label{equ:update-sigma}
\end{split}
\end{equation}
%The complete gradient derivations and Kronecker product computations are provided in Appendix \ref{appx:m-step-details}. 
Combining the estimation for the first and second group, the detailed algorithm is shown in Appendix \ref{appx:m-step-details}.

\subsection{Parameter selection}\label{sec:parameter-selection}
We now provide a guideline for selecting the number of latent factors $m$, which is a key hyperparameter in the proposed model. To this end, we adopt the Akaike Information Criterion (AIC), a widely used criterion for evaluating the relative quality of statistical models. The AIC is defined as ${\rm AIC} = - 2 \log p(\bX|\hat{\Theta}) + 2 M$, where $\hat{\Theta}$ denotes the estimated model parameters, and  $M$ is the total number of free parameters in the model. However, in our setting, the marginal likelihood $\log p(\bX|\hat{\Theta})$ does not have a closed-form solution due to the latent variables $\bZ$. Therefore, we approximate it using the variational lower bound by replacing $\log p(\bX|\hat{\Theta})$ with $\underline{\bbE_q(\log p(\bX | \bZ, \hat{\Theta}))}$, and get 
\begin{equation} \label{equ:AIC}
    {\rm AIC} = - 2 \underline{\bbE_q(\log p(\bX | \bZ, \hat{\Theta}))} + 2 (nm + 2K^2 + 2m^2 + n^2K + m^2K + 2).
\end{equation}
Here $\hat{\Theta}$ is the final estimated parameters according to Algorithm \ref{alg:EM} and $q(\bZ)$ in $\underline{\bbE_q(\log p(\bX | \bZ, \hat{\Theta}))}$ is the corresponding approximation for $p(\bZ|\bX,\hat{\Theta})$ computed in the final E-step. To find the appropriate $m$, we can set an upper limit $m_{\max}$ for $m$ and fit the model separately for each $m$ from $1$ to $m_{\max}$. Then we calculate the AIC according to Eq. (\ref{equ:AIC}) and finally find the best $m$ with the smallest AIC.

\section{Numerical Studies}
\label{sec:experiments}
In this section, we evaluate the performance of TSSDMN in terms of both estimation and prediction errors under various experiment settings. To benchmark our method, we compare it against the following four baseline models: (1) \textbf{EDMN:} Eigenmodel for dynamic multilayer networks proposed by \cite{loyal2023eigenmodel}, which captures shared latent dynamics using a random walk framework; (2) \textbf{MTR:} Multilinear tensor regression proposed by \cite{hoff2015multilinear}, a tensor-based regression model for multilayer relational data; (3) \textbf{BDMN:} Bayesian dynamic multilayer network proposed by \cite{durante2017bayesian}, which models latent factors via Gaussian processes across time and layers; (4) \textbf{DSBM:} Dynamic stochastic block model proposed by \cite{xu2014dynamic}, designed for single-layer dynamic networks. To apply it in the multilayer setting, we estimate the model independently for each layer. % \chen{explain more for each baseline}

To investigate the impact of different model configurations, we conduct three experiments that separately vary the number of nodes, the number of layers, and the variance of latent variables. The experimental settings are summarized as follows:
    \begin{itemize}
        \item Setting 1: Number of nodes $n \in [10, 20, 30, 40, 50]$, number of layers $K=2$, number of time steps $T=30$, variance parameters $\sigma^2 = \omega^2= 0.01$.
        \item Setting 2: Number of nodes $n = 20$,number of layers $K \in [1, 2, 3, 4, 5]$, number of time steps $T=30$, variance parameters $\sigma^2 = \omega^2= 0.01$.
        \item Setting 3: Number of nodes $n = 20$, number of layers $K$ = 2, number of time steps $T=30$, variance parameters $\sigma^2=\omega^2 \in [0.01, 0.04, 0.09, 0.25]$.
    \end{itemize}
To assess the efficiency and robustness of these models under scenarios where the data-generating process deviates from the model assumptions, we conduct simulations using three different generative mechanisms: TSSDMN, EDMN and BDMN. For each experimental setting described above, we generate dynamic multilayer network data from each of these three models and assess the performance of all five candidate models (TSSDMN, EDMN, BDMN, MTR, and DSBM) on each generated dataset.
Note that MTR is specifically designed for continuous-valued data and DSBM is a model for single-layer networks. Therefore, we exclude these two models from the data-generating processes, though we still include them as competing methods during evaluation.

The data generation procedure for TSSDMN is as follows: (1) Generate positive definite matrices $\mathbf{A}_i$ with $\|\mathbf{A}_i\|_2 < 1$ for $i=1,2,3$; Generate latent node factors $\bC_1$ with nonnegative entries and a layer-transition matrix $\bC_2$; (2) Sample the initial latent mean vector $\mathbf{u}_0$ from a uniform distribution $\mathcal{U}(-1, 1)$; (3) Generate the the latent interaction tensors as follows: $\mathbf{z}_0 \sim \mathcal{N}(\mathbf{u}_0, \omega^2)$, $\mathbf{z}_t \sim \mathcal{N}(\mathbf{A} \mathbf{z}_{t-1}, \sigma^2 \mathbf{I})$ for all $t \geq 1$ where $\bA = \bA_1 \otimes \bA_2 \otimes \bA_3$; (4) Compute the log-odds of edge formulation as $\boldsymbol{\gamma}_t = \mathbf{C} \mathbf{z}_t$, and generate observations via $\mathbf{x}_{t,i} \sim \text{Bernoulli}\left(\frac{\exp(\boldsymbol{\gamma}_{t,i})}{1 + \exp(\boldsymbol{\gamma}_{t,i})}\right)$ for all $t \geq 1$. The data generation procedures for EDMN and BDMN follow their respective original formulations and are detailed in Appendix. \ref{app:data-generation}. When fitting TSSDMN to data generated from other models, we use our proposed AIC in Section \ref{sec:parameter-selection} to select the best $m$. 
    \begin{figure}[t]
        \centering
        \includegraphics[width = \linewidth]{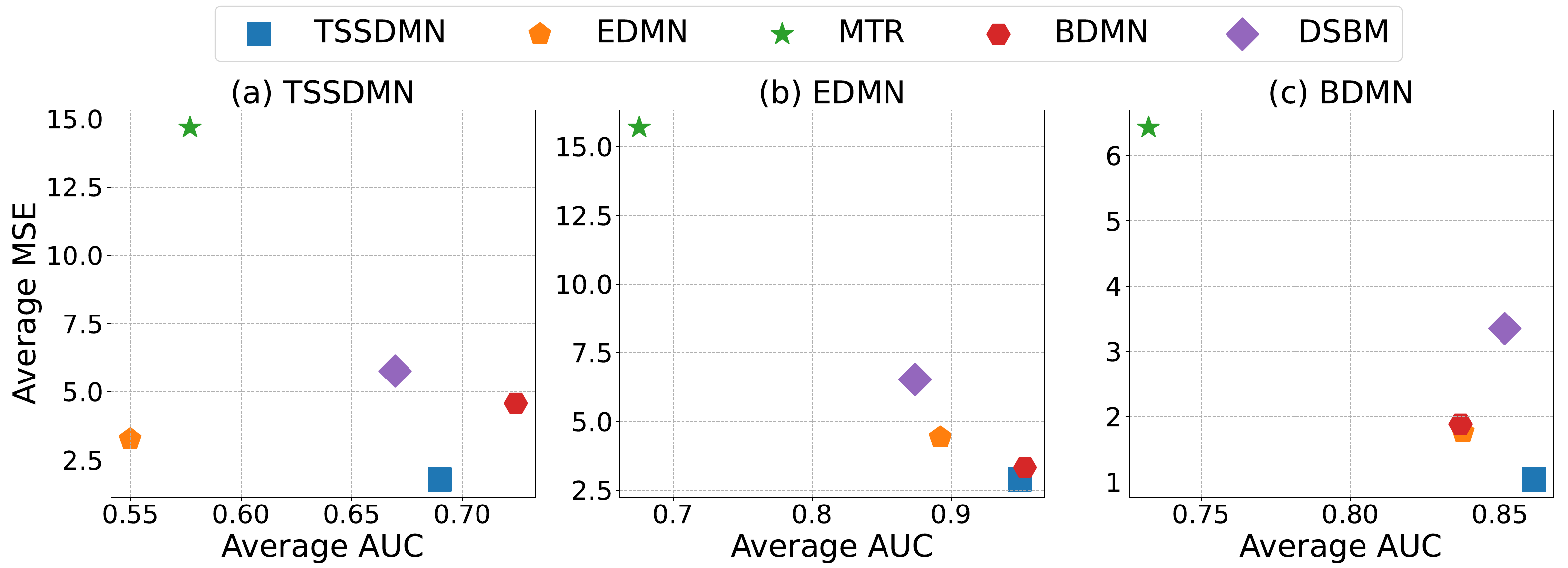}
        \caption{Average AUC and MSE of different models when the data is generated from (a) TSSDMN, (b) EDMN, (c) BDMN.
        \label{fig:result_pos}}
    \end{figure}

\begin{table}[t]
\centering
\resizebox{\columnwidth}{!}{
\begin{tabular}{ccccccc}
\toprule
Datasets                     & Nodes & \textbf{TSSDMN}      & \textbf{EDMN}        & \textbf{MTR} & \textbf{BDMN}        & \textbf{DSBM} \\
\toprule
\multirow{5}{*}{\textbf{TSSDMN}} & 10    & \textbf{3.90(±0.18)} & \underline{7.98(±0.25)}          & 16.56(±1.38) & 8.94(±0.20)          & 12.16(±0.55)  \\
                                 & 20    & \textbf{1.73(±0.06)} & \underline{3.78(±0.20)}          & 6.42(±0.19)  & 5.18(±0.11)          & 5.89(±0.10)   \\
                                 & 30    & \textbf{1.16(±0.03)} & \underline{2.25(±0.10)}          & 4.66(±0.10)  & 3.67(±0.10)          & 4.00(±0.06)   \\
                                 & 40    & \textbf{1.08(±0.02)} & \underline{1.36(±0.06)}          & 3.54(±0.06)  & 2.83(±0.06)          & 2.83(±0.05)   \\
                                 & 50    & \underline{1.12(±0.06)}          & \textbf{0.99(±0.06)} & 3.38(±0.06)  & 2.28(±0.04)          & 2.17(±0.04)   \\
\cmidrule{2-7}
\multirow{5}{*}{\textbf{EDMN}}   & 10    & \textbf{5.28(±0.42)} & \underline{5.68(±0.93)}          & 26.50(±2.00) & 5.92(±0.59)          & 10.62(±0.52)  \\
                                 & 20    & \textbf{3.32(±0.17)} & 4.14(±0.65)          & 17.55(±1.06) & \underline{3.90(±0.24)}          & 7.09(±0.39)   \\
                                 & 30    & \textbf{2.29(±0.13)} & 4.03(±0.83)          & 12.33(±0.66) & \underline{2.82(±0.19)}          & 5.34(±0.19)   \\
                                 & 40    & \textbf{1.95(±0.11)} & 3.53(±0.79)          & 11.56(±0.83) & \underline{2.24(±0.15)}          & 4.68(±0.16)   \\
                                 & 50    & \textbf{1.61(±0.09)} & 4.71(±0.86)          & 10.58(±0.95) & \underline{1.72(±0.13)}          & 4.90(±0.13)   \\
\cmidrule{2-7}
\multirow{5}{*}{\textbf{BDMN}}   & 10    & \textbf{1.45(±0.23)} & 4.65(±0.08)          & 11.76(±0.67) & \underline{3.03(±0.05)}          & 8.74(±0.42)   \\
                                 & 20    & \textbf{0.89(±0.02)} & \underline{1.70(±0.03)}          & 9.22(±0.91)  & 2.12(±0.03)          & 3.41(±0.09)   \\
                                 & 30    & \textbf{0.85(±0.02)} & \underline{1.05(±0.03)}          & 3.87(±0.11)  & 1.64(±0.03)          & 1.98(±0.04)   \\
                                 & 40    & \underline{0.83(±0.02)}          & \textbf{0.75(±0.02)} & 3.41(±0.28)  & 1.36(±0.03)          & 1.41(±0.03)   \\
                                 & 50    & \underline{1.18(±0.16)}          & \textbf{0.68(±0.03)} & 3.89(±0.37)  & 1.28(±0.03)          & 1.21(±0.03)   \\
\bottomrule
\end{tabular}}
\caption{Average MSE of different models for data generated from different scenarios in Setting 1 (with standard deviations in parathesis)
\label{tab:MSE-node}}
\end{table}
\begin{table}[t]
% \usepackage{multirow}
\centering
\resizebox{\columnwidth}{!}{
\begin{tabular}{ccccccc}
\toprule
Datasets           & Nodes & \textbf{TSSDMN}        & \textbf{EDMN}          & \textbf{MTR}           & \textbf{BDMN}          & \textbf{DSBM}          \\
\toprule
\multirow{5}{*}{\textbf{TSSDMN}} & 10    & \underline{75.7(±0.8)}  & 58.0(±0.5)  & 55.3(±0.4)  & \textbf{82.3(±0.6)}  & 71.7(±1.2)  \\
                        & 20    & \underline{71.2(±0.6)} & 55.6(±0.3) & 55.9(±0.1) & \textbf{73.9(±0.6)} & 70.7(±0.4) \\
                        & 30    & \underline{68.6(±0.4)}  & 54.4(±0.2)  & 56.2(±0.1)  & \textbf{70.5(±0.3)}  & 66.3(±0.2)  \\
                        & 40    & \underline{65.7(±0.3)}  & 53.6(±0.1)  & 56.4(±0.1)  & \textbf{68.1(±0.2)}  & 63.7(±0.2)  \\
                        & 50    & \underline{63.7(±0.4)}  & 53.3(±0.1)  & 56.4(±0.1)  & \textbf{67.3(±0.2)}  & 62.4(±0.2)  \\
\cmidrule{2-7}
\multirow{5}{*}{\textbf{EDMN}}   & 10    & \underline{95.4(±0.6)}  & 89.6(±1.7)  & 54.9(±0.7)  & \textbf{96.0(±0.6)}  & 86.7(±0.9)  \\
                        & 20    & \underline{94.8(±0.6)}  & 90.3(±1.2)  & 63.2(±1.6)  & \textbf{95.2(±0.6)}  & 85.9(±1.0)  \\
                        & 30    & \underline{94.9(±0.6)}  & 89.2(±1.9)  & 71.7(±2.1)  & \textbf{95.2(±0.6)}  & 88.2(±0.6)  \\
                        & 40    & \underline{94.7(±0.4)}  & 89.5(±1.8)  & 71.7(±2.4)  & \textbf{95.0(±0.4)}  & 88.2(±0.5)  \\
                        & 50    & \underline{94.9(±0.4)}  & 87.5(±2.0)  & 74.8(±2.7)  & \textbf{95.2(±0.5)}  & 88.1(±0.5)  \\
\cmidrule{2-7}
\multirow{5}{*}{\textbf{BDMN}}   & 10    & \textbf{88.0(±0.6)}  & 85.9(±0.2)  & 70.7(±0.7)  & \underline{86.1(±0.2)}  & 85.2(±1.0)  \\
                        & 20    & \textbf{87.1(±0.2)}  & 84.3(±0.2)  & 69.6(±1.7)  & 84.4(±0.1)  & \underline{86.4(±0.3)}  \\
                        & 30    & \textbf{86.2(±0.2)}  & 83.5(±0.3)  & 77.0(±0.4)  & 83.5(±0.2)  & \underline{85.8(±0.2)}  \\
                        & 40    & \textbf{85.8(±0.2)}  & 82.9(±0.2)  & 76.1(±1.0)  & 82.7(±0.2)  & \underline{84.8(±0.2)}  \\
                        & 50    & \textbf{83.6(±0.8)}  & 82.2(±0.3)  & 73.6(±1.2)  & 81.7(±0.2)  & \textbf{83.6(±0.2)}  \\
\bottomrule
\end{tabular}
}
\caption{Average AUC of different models for data generated from different scenarios in Setting 1 (with standard deviations in parentheses) 
\label{tab:AUC-node}}
\end{table}

We evaluate model performance using two standard metrics:
Mean Square Error (MSE) and the Area Under the Curve (AUC), defined as follows:
     \begin{align*}
     {\rm MSE} &= \frac{1}{Tn^2K} \sum_{t=1}^T \| \hat{\bgamma}_t - \bgamma_t \|_2^2, \\
     {\rm AUC} &= \frac{1}{n_0 n_1} \sum_{\mathbb{I}(\calX_{t,i'j'k'}=1)} \sum_{\mathbb{I}(\calX_{t,i'j'k'}=0)} \mathbb{I}(\bGamma_{t,ijk} > \bGamma_{t,i'j'k'}), \label{equ:auc}
     \end{align*}
where $n_0 = \sum_{\mathbb{I}(\calX_{t,i'j'k'}=0)}1$, $n_1 = \sum_{\mathbb{I}(\calX_{t,i'j'k'}=1)} 1$ are the number of non-existing and existing edges,  and $\mathbb{I}(\cdot)$ is the indicator function. The MSE measures the discrepancy between the estimated and true latent log-odds, reflecting how accurately the model recovers the underlying latent structure. The AUC, on the other hand, assesses the model's ability to discriminate between connected and non-connected node pairs based on their predicted scores, indicating how well the model fits the observed data. It is important to interpret MSE and AUC jointly, which enables a more nuanced understanding of both inference quality and predictive performance of a model. In particular, low MSE and high AUC  suggest that the model not only fits the true latent structure well but also aligns with the observed network data. High AUC but high MSE may indicate overfitting, where the model captures the observed edges well but fails to generalize to the underlying data-generating mechanism. Low MSE but low AUC may reflect high variance, where the model captures the latent structure but performs poorly in edge prediction.
High MSE and low AUC suggest a generally poor model fit 

Figure~\ref{fig:result_pos} reports the average AUC and MSE of the five models, evaluated over 30 independent experiment replicates under the three data generation scenarios in Setting 1. Overall, TSSDMN consistently achieves the lowest or near-lowest MSE across all scenarios, demonstrating superior accuracy in estimating the latent connection probabilities. Although TSSDMN does not always yield the highest AUC, its AUC values remain competitive and stable. In contrast, for the other four models, though they occasionally achieve higher AUCs than TSSDMN, their MSEs are quite high, suggesting significant overfitting. These results highlight the effectiveness and robustness of TSSDMN, which strikes a favorable balance between predictive performance and model generalization, reliably capturing the latent dynamics without overfitting.

\begin{table}[t]
\centering
\resizebox{\columnwidth}{!}{
\begin{tabular}{ccccccc}
\toprule
Datasets                    & Layers & \textbf{TSSDMN}      & \textbf{EDMN}        & \textbf{MTR} & \textbf{BDMN} & \textbf{DSBM} \\
\toprule
\multirow{5}{*}{\textbf{TSSDMN}} & 1      & \textbf{3.27(±0.22)} & \underline{8.15(±0.46)}          & 30.63(±0.38) & 8.50(±0.25)   & 11.41(±0.61)  \\
                                 & 2      & \textbf{3.90(±0.18)} &  \underline{7.98(±0.25)}          & 16.56(±1.38) & 8.94(±0.20)   & 12.16(±0.55)  \\
                                 & 3      & \textbf{5.58(±0.26)} &  \underline{7.82(±0.29)}          & 10.84(±0.38) & 10.41(±0.31)  & 11.51(±0.32)  \\
                                 & 4      & \textbf{4.33(±0.14)} &  \underline{7.76(±0.14)}          & 9.89(±0.18)  & 10.15(±0.28)  & 12.27(±0.34)  \\
                                 & 5      & \textbf{4.41(±0.13)} &  \underline{7.32(±0.20)}          & 8.69(±0.19)  & 10.64(±0.24)  & 11.62(±0.27)  \\
\cmidrule{2-7}
\multirow{5}{*}{\textbf{EDMN}}   & 1      &  \underline{5.42(±0.54)}          & \textbf{5.06(±0.77)} & 39.16(±0.87) & 5.90(±0.55)   & 10.93(±0.76)  \\
                                 & 2      & \textbf{5.28(±0.42)} &  \underline{5.68(±0.93)}          & 26.50(±2.00) & 5.92(±0.59)   & 10.62(±0.52)  \\
                                 & 3      &  \underline{5.28(±0.32)}          & \textbf{4.65(±0.84)} & 20.22(±1.01) & 6.71(±0.53)   & 10.20(±0.27)  \\
                                 & 4      &  \underline{5.21(±0.30)}          & \textbf{3.70(±0.80)} & 18.26(±0.55) & 6.50(±0.60)   & 10.86(±0.50)  \\
                                 & 5      &  \underline{5.12(±0.24)}          & \textbf{4.23(±0.84)} & 17.23(±0.54) & 6.78(±0.52)   & 10.57(±0.27)  \\
\cmidrule{2-7}
\multirow{5}{*}{\textbf{BDMN}}   & 1      & \textbf{1.05(±0.22)} & 4.94(±0.10)          & 33.32(±0.90) &  \underline{2.49(±0.05)}   & 8.39(±0.35)   \\
                                 & 2      & \textbf{1.45(±0.23)} & 4.65(±0.08)          & 11.76(±0.67) &  \underline{3.03(±0.05)}   & 8.74(±0.42)   \\
                                 & 3      & \textbf{2.52(±0.44)} & 4.57(±0.07)          & 9.60(±0.33)  &  \underline{2.87(±0.04)}   & 8.31(±0.18)   \\
                                 & 4      & \textbf{0.86(±0.02)} & 4.46(±0.06)          & 7.43(±0.10)  &  \underline{2.75(±0.05)}   & 8.48(±0.15)   \\
                                 & 5      & \textbf{0.85(±0.01)} & 4.55(±0.08)          & 6.72(±0.15)  &  \underline{2.68(±0.05)}   & 8.74(±0.20)  \\
\bottomrule
\end{tabular}
}
\caption{Average MSE of different models for data generated from different scenarios in Setting 2 (with standard deviations in parathesis) 
\label{tab:MSE-layer}}

\end{table}

\begin{table}[t]
\resizebox{\columnwidth}{!}{
\begin{tabular}{ccccccc}
\toprule
Datasets                  & Layers & \textbf{TSSDMN} & \textbf{EDMN} & \textbf{MTR}  & \textbf{BDMN} & \textbf{DSBM} \\
\toprule
\multirow{5}{*}{\textbf{TSSDMN}} & 1      & \underline{77.3(±0.9)}   & 58.8(±1.0)  & 56.4(±0.4)  & \textbf{78.2(±1.0)}  & 74.4(±1.3)  \\
                                 & 2      & \underline{75.7(±0.8)}   & 58.0(±0.5)  & 55.3(±0.4)  & \textbf{82.3(±0.6)}  & 71.7(±1.2)  \\
                                 & 3      & 70.2(±1.1)   & 57.3(±0.4)  & 53.5(±0.2)  & \textbf{78.7(±0.8)}  & \underline{74.1(±0.8)}  \\
                                 & 4      & \underline{75.0(±0.5)}   & 57.6(±0.6)  & 53.2(±0.2)  & \textbf{77.6(±0.5)}  & 72.8(±0.7)  \\
                                 & 5      & \underline{74.0(±0.6)}   & 57.5(±0.4)  & 53.2(±0.2)  & \textbf{75.3(±0.4)}  & 73.8(±0.6)  \\
\cmidrule{2-7}
\multirow{5}{*}{\textbf{EDMN}}   & 1      & \textbf{95.8(±0.6)}   & 93.2(±0.8)  & 56.6(±0.3)  & \underline{95.6(±0.7)}  & 85.8(±1.4)  \\
                                 & 2      & \underline{95.4(±0.6)}   & 89.6(±1.7)  & 54.9(±0.7)  & \textbf{96.0(±0.6)}  & 86.7(±0.9)  \\
                                 & 3      & \underline{94.5(±0.6)}   & 89.9(±1.4)  & 54.3(±0.9)  & \textbf{94.9(±0.6)}  & 86.8(±0.5)  \\
                                 & 4      & \underline{94.8(±0.5)}   & 91.4(±1.3)  & 53.4(±0.3)  & \textbf{95.1(±0.5)}  & 85.8(±0.8)  \\
                                 & 5      & \textbf{94.8(±0.6)}   & 90.4(±1.3)  & 53.4(±0.3)  & \underline{94.7(±0.6)}  & 86.5(±0.5)  \\
\cmidrule{2-7}
\multirow{5}{*}{\textbf{BDMN}}   & 1      & \textbf{90.7(±0.5)}   & \underline{87.8(±0.4)} & 55.8(±0.2)  & 84.1(±0.3)  & 87.0(±0.9)  \\
                                 & 2      & \textbf{88.0(±0.6)}   & 85.9(±0.2)  & 70.7(±0.7)  & \underline{86.1(±0.2)}  & 85.2(±1.0)  \\
                                 & 3      & \textbf{85.8(±0.9)}   & \underline{86.0(±0.2)} & 72.3(±0.6)  & 85.2(±0.2)  & 85.7(±0.6)  \\
                                 & 4      & \textbf{88.7(±0.2)}   & \underline{85.6(±0.2)} & 74.5(±0.3)  & 85.1(±0.2)  & 84.6(±0.6)  \\
                                 & 5      & \textbf{89.2(±0.2)}   & \underline{86.3(±0.2)} & 76.5(±0.4)  & 85.1(±0.2)  & 84.1(±0.7)  \\
\bottomrule
\end{tabular}
}
\caption{Average AUC of different models for data generated from different scenarios in Setting 2 (with standard deviations in parentheses)
\label{tab:AUC-layer}}
\end{table}

\begin{table}[t]
\centering
\resizebox{\columnwidth}{!}{
\begin{tabular}{ccccccc}
\toprule
Datasets                   & Variance & \textbf{TSSDMN}      & \textbf{EDMN}        & \textbf{MTR} & \textbf{BDMN} & \textbf{DSBM} \\
\toprule
\multirow{4}{*}{\textbf{TSSDMN}} & 0.1      & \textbf{3.90(±0.18)} &  \underline{7.98(±0.25)}          & 16.56(±1.38) & 8.94(±0.20)   & 12.16(±0.55)  \\
                                 & 0.2      & \textbf{4.89(±0.29)} & 13.50(±0.45)         & 19.65(±1.10) &  \underline{9.48(±0.32)}   & 12.89(±0.48)  \\
                                 & 0.3      & \textbf{4.79(±0.21)} & 17.14(±0.45)         & 22.64(±1.20) &  \underline{10.16(±0.19)}  & 13.63(±0.68)  \\
                                 & 0.5      & \textbf{5.08(±0.25)} & 20.10(±0.28)         & 25.86(±1.36) &  \underline{9.65(±0.29)}   & 13.66(±0.57)  \\
\cmidrule{2-7}
\multirow{4}{*}{\textbf{EDMN}}   & 0.1      & \textbf{5.28(±0.42)} &  \underline{5.68(±0.93)}          & 26.50(±2.00) & 5.92(±0.59)   & 10.62(±0.52)  \\
                                 & 0.2      & \textbf{5.25(±0.33)} &  \underline{5.43(±0.77)}          & 24.14(±1.57) & 5.48(±0.49)   & 10.80(±0.54)  \\
                                 & 0.3      &  \underline{6.23(±0.40)}          & \textbf{6.07(±0.83)} & 27.67(±1.98) & 6.43(±0.49)   & 10.74(±0.63)  \\
                                 & 0.5      & 8.96(±0.48)          & \textbf{4.77(±0.73)} & 26.12(±1.28) &  \underline{8.38(±0.62)}   & 10.08(±0.44)  \\
\cmidrule{2-7}
\multirow{4}{*}{\textbf{BDMN}}   & 0.1      & \textbf{1.45(±0.23)} & 4.65(±0.08)          & 11.76(±0.67) & 3.03(±0.05)   & 8.74(±0.42)   \\
                                 & 0.2      & \textbf{1.50(±0.26)} & 4.55(±0.08)          & 11.71(±0.59) &  \underline{3.00(±0.04)}   & 8.14(±0.17)   \\
                                 & 0.3      & \textbf{1.61(±0.30)} & 4.66(±0.08)          & 12.30(±0.91) &  \underline{2.98(±0.05)}   & 8.18(±0.20)   \\
                                 & 0.5      & \textbf{1.58(±0.28)} & 4.59(±0.08)          & 11.31(±0.57) &  \underline{3.04(±0.05)}   & 8.55(±0.29)   \\
% \cmidrule{2-7}
% \multirow{4}{*}{\textbf{DSBM}}   & 0.1      & \textbf{1.82(±0.11)} & 5.02(±0.76)          & 22.11(±1.30) &  \underline{2.27(±0.09)}   & 6.62(±0.62)   \\
%                                  & 0.2      & \textbf{2.20(±0.11)} & 5.16(±0.31)          & 23.58(±1.58) &  \underline{2.66(±0.08)}   & 6.47(±0.59)   \\
%                                  & 0.3      & \textbf{2.50(±0.16)} & 6.45(±0.60)          & 27.32(±1.93) &  \underline{3.20(±0.14)}   & 5.44(±0.25)   \\
%                                  & 0.5      & \textbf{3.75(±0.31)} & 8.48(±0.69)          & 29.67(±2.82) & 7.31(±0.46)   &  \underline{5.54(±0.77)}  \\
\bottomrule
\end{tabular}
}
\caption{Average MSE of different models for data generated from different scenarios in Setting 3 (with standard deviations in parathesis)
\label{tab:MSE-sigma}}
\end{table}

\begin{table}[t]
\centering
\resizebox{\columnwidth}{!}{
\begin{tabular}{ccccccc}
\toprule
\multicolumn{1}{l}{Datasets}   & \multicolumn{1}{l}{Variance} & \textbf{TSSDMN} & \textbf{EDMN} & \textbf{MTR}  & \textbf{BDMN} & \textbf{DSBM} \\
\toprule
\multirow{4}{*}{\textbf{TSSDMN}} & 0.1                          & \underline{75.7(±0.8)}   & 58.0(±0.5)  & 55.3(±0.4)  & \textbf{82.3(±0.6)}  & 71.7(±1.2)  \\
                                 & 0.2                          & \underline{84.6(±1.0)}   & 59.1(±0.8)  & 55.3(±0.8)  & \textbf{89.9(±0.6)}  & 76.8(±1.2)  \\
                                 & 0.3                          & \underline{89.8(±0.6)}   & 61.3(±1.1)  & 54.8(±0.7)  & \textbf{93.1(±0.3)}  & 79.7(±1.4)  \\
                                 & 0.5                          & \underline{93.4(±0.6)}   & 64.0(±1.1)  & 54.7(±0.8)  & \textbf{96.0(±0.3)}  & 81.8(±1.3)  \\
\cmidrule{2-7}
\multirow{4}{*}{\textbf{EDMN}}   & 0.1                          & \underline{95.4(±0.6)}   & 89.6(±1.7)  & 54.9(±0.7)  & \textbf{96.0(±0.6)}  & 86.7(±0.9)  \\
                                 & 0.2                          & \underline{94.2(±0.6)}   & 90.0(±1.4)  & 55.8(±1.0)  & \textbf{94.9(±0.7)}  & 87.5(±0.8)  \\
                                 & 0.3                          & \underline{94.5(±0.5)}   & 90.7(±1.2)  & 56.1(±1.4)  & \textbf{95.6(±0.5)}  & 86.7(±0.9)  \\
                                 & 0.5                          & \underline{94.0(±0.6)}   & 93.9(±0.8)  & 56.8(±1.7)  & \textbf{95.8(±0.4)}  & 87.6(±0.6)  \\
\cmidrule{2-7}
\multirow{4}{*}{\textbf{BDMN}}   & 0.1                          & \textbf{88.0(±0.6)}   & 85.9(±0.2)  & 70.7(±0.7)  & \underline{86.1(±0.2)}  & 85.2(±1.0)  \\
                                 & 0.2                          & \textbf{87.6(±0.6)}   & 85.9(±0.3)  & 70.4(±0.7)  & 85.7(±0.2)  & \underline{86.3(±0.5)}  \\
                                 & 0.3                          & \textbf{87.6(±0.6)}   & 85.8(±0.2)  & 70.1(±0.8)  & \underline{86.2(±0.2)}  & 86.0(±0.7)  \\
                                 & 0.5                          & \textbf{87.5(±0.7)}   & \underline{86.0(±0.2)} & 71.0(±0.6)  & 85.7(±0.2)  & 85.2(±0.8)  \\
\bottomrule
\end{tabular}
}
\caption{Average AUC of different models for data generated from different scenarios in Setting 3 (with standard deviations in parentheses)
\label{tab:AUC-sigma}}
\end{table}

\begin{figure}[h]
        \centering
        \includegraphics[width = \linewidth]{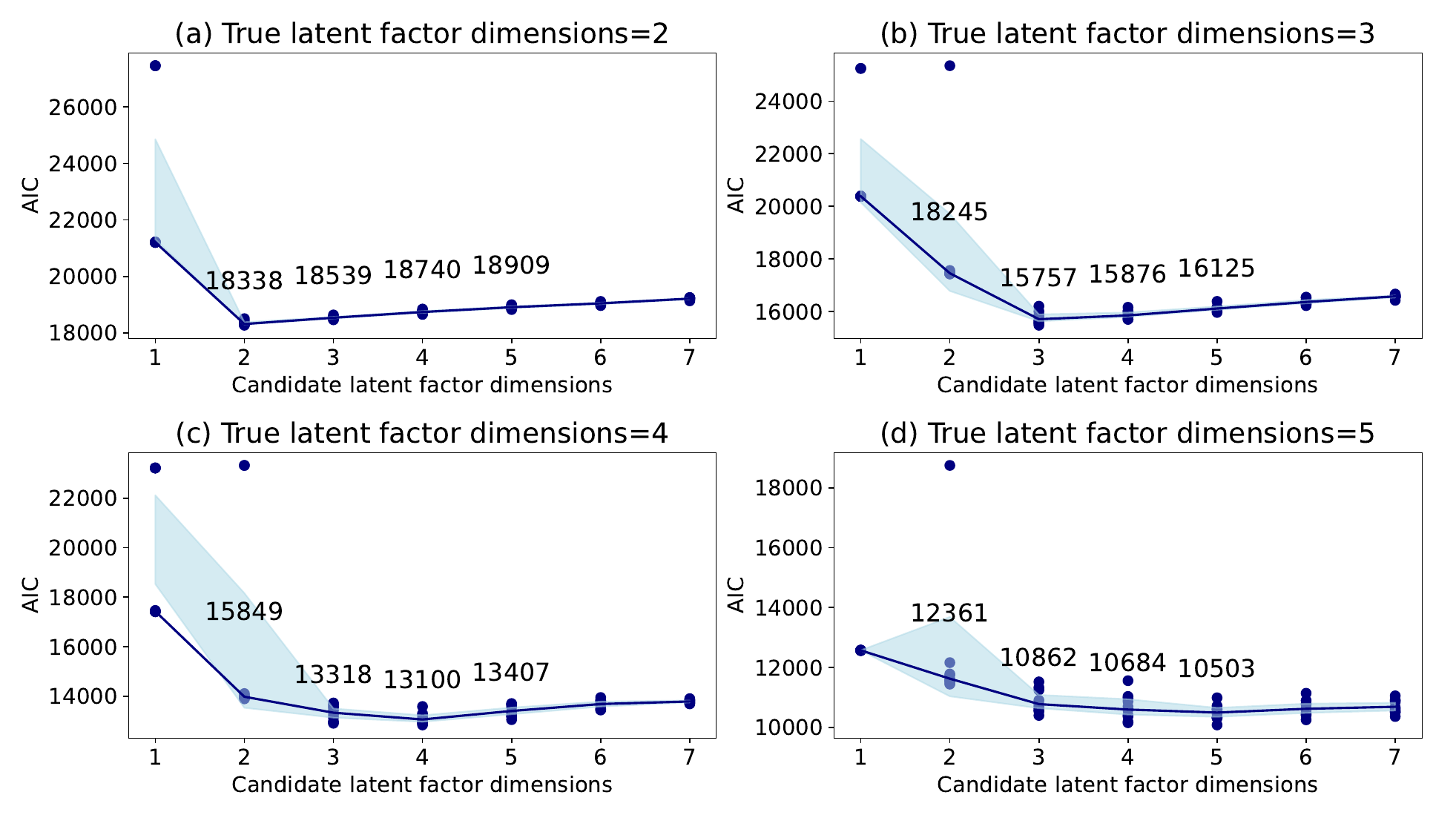}
        \caption{AIC plot for TSSDMN with the true latent factor dimensions (a) $m=2$, (b) $m=3$, (c) $m=4$, (d) $m=5$.
        \label{fig:AIC-numerical-study}}
    \vspace{-10pt}
\end{figure}

Tables \ref{tab:MSE-node} to \ref{tab:AUC-sigma} present the detailed performance metrics (MSE and AUC) for all five models across the different data generation scenarios in the three experimental settings. TSSDMN consistently achieves either the best or second-best performance in both MSE and AUC across all scenarios, demonstrating its robustness and reliability under varying conditions. In addition, as the number of nodes increases, the performance of all the models increases. This can be attributed to the fact that a larger number of nodes increases both the number of model parameters and the amount of observed data. However, the growth in data volume outpaces the growth in parameter complexity, leading to better estimation and filtering performance. In contrast, increasing the number of layers results in only marginal changes in model performance. This is likely because both the number of model parameters and the amount of data increase at similar rates, keeping the effective signal-to-noise ratio relatively stable. Last, all models exhibit larger MSEs, reflecting the greater difficulty of recovering accurate latent representations from noisier data. Nevertheless, TSSDMN remains among the top-performing models in terms of MSE. Interestingly, increasing the variance also leads to higher AUC values across all models. This can be explained by the tendency of the Bernoulli probabilities to become more extreme (closer to 0 or 1) as variance increases, which in turn improves the separability between positive and negative edges, resulting in elevated AUC scores.

We further evaluate the effectiveness of the latent factor dimension selection algorithm based on AIC introduced in Section~\ref{sec:parameter-selection}. We set the basic experiment settings as follows:
\begin{itemize}
\item Setting 4: Number of nodes $n=20$, number of layers $K=2$, number of time steps $T=30$, true latent dimension $m\in [2,3,4,5]$, and noise variances $\sigma^2 =\omega^2= 0.01$.
\end{itemize}
For each scenario, we consider a candidate latent factor dimension from 1 and 7, and calculate the AIC for each candidate. Each experiment is replicated 10 times, and the average AIC values for each candidate dimension across all scenarios are shown in Figure~\ref{fig:AIC-numerical-study}.We can see that the minimum AIC is always achieved when the candidate latent factor dimension matches the true one, demonstrating that the proposed AIC-based selection method is both accurate and reliable in identifying the appropriate number of latent factors.

\begin{figure}[t]
    \centering
    \includegraphics[width = \linewidth]{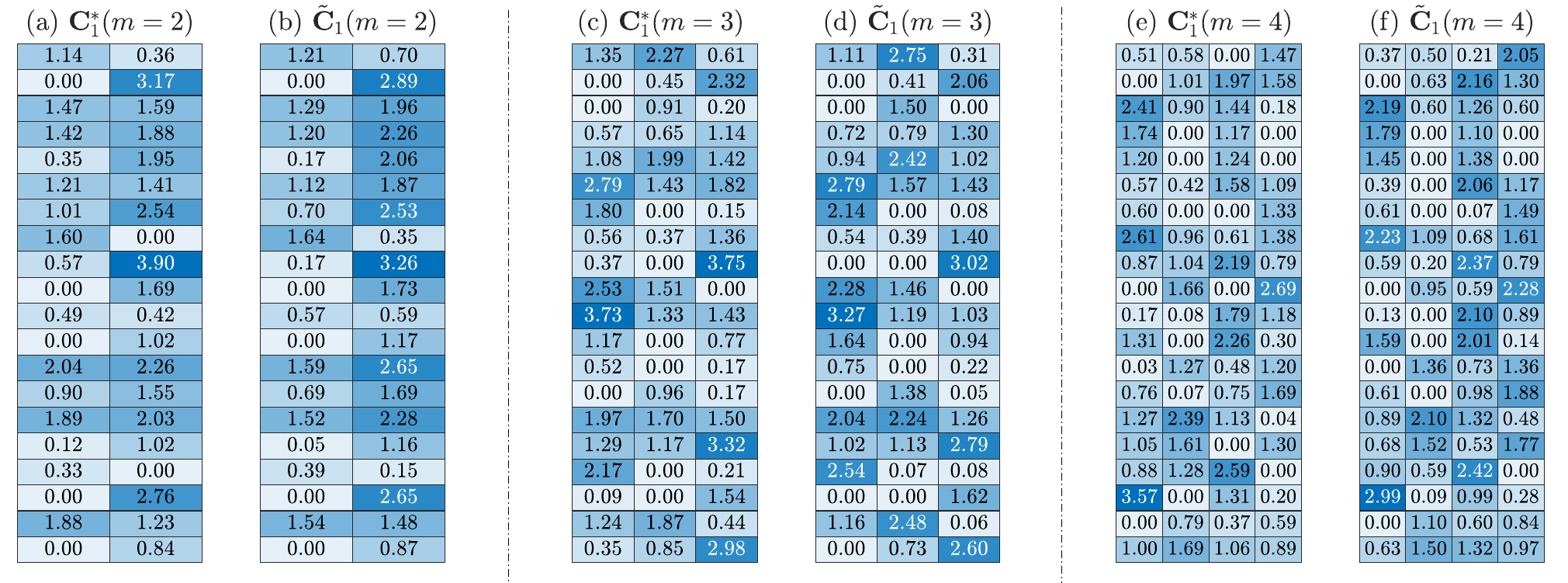}
    \caption{Comparison between $\bC_1^*$ and $\tilde{\bC}_1$ with different $m$.
    \label{fig:Community-detection}}
\end{figure}

Last, we evaluate whether the true latent factors can be identifiable up to a permutation matrix, as suggested in Theorem \ref{the:identifiability}. However, in our experiment, the estimated factors $\hat{\bC}_1, \hat{\bC}_2$ and the true factors $\bC_1^*, \bC_2^*$ do not fully satisfy assumption A1. Therefore, identifiability is only guaranteed up to a permutation matrix and a global scaling constant. So we test the estimation accuracy of the latent factor by comparing the true latent factor matrix $\bC_1^*$ against an adjusted latent factor $\tilde{\bC}_1$, obtained by aligning $\hat{\bC}_1$ via $ \tilde{\bC}_1 = \underset{\bC_1 \in \{c \bPi \hat{\bC}_1 | \forall c, \bPi\}}{\arg \min} \| \bC_1^* - \bC_1 \|_F^2$,  where $r$ is a scalar and $\bPi$ is a permutation matrix. Figure \ref{fig:Community-detection} illustrates the true latent matrix $\bC_1^*$ and the estimated latent factor after rotation, i.e.,  $\tilde{\bC}_1$, under different latent factor dimensions. The visual similarity between the two matrices confirms that the underlying structure is well captured by the estimated latent factors. To quantify this, we compute the Mean Absolute Percentage Error (MAPE) between the true and adjusted matrices: ${\rm MAPE}=\frac{\left\|\bC_1^*-\tilde{\bC}_1 \right\|_F}{\left\| \bC_1^* \right\|_F}$. Over 10 independent replications, the average MAPE is 0.159 with a standard deviation of 0.020, which indicates a robust estimation of the latent factors.

\section{Case Studies}
\label{sec:case}
\subsection{ICEWS coded event data}
\label{sec:case1}
The Integrated Crisis Early Warning System (ICEWS) dataset provides machine-coded records of political and socio-economic interactions among countries dating back to January 1991 \cite{boschee2015icews}. Each event in the dataset is represented as a quadruple: (source country, target country, event type, timestamp). The event types are categorized into four classes—verbal cooperation, material cooperation, verbal conflict, and material conflict—which range from the most cooperative to the most adversarial interactions. This dataset has been widely used for modeling multilayer dynamic networks, including in recent work such as \cite{loyal2023eigenmodel}.

% For our analysis, we focus on the 30 most prominent countries and examine their interactions over the period from 2009 to 2017. Each country is treated as a node, and each event type corresponds to a distinct layer. An event is represented as a directed edge from the source country to the target country in the corresponding layer. This formulation results in a dynamic multilayer network with $n=30$ nodes, $T=95$ time steps (one per month from January 2009), and $K=4$ layers (corresponding to the four event types). We define $\bcalX_{t,ijk}=1$ if at least one event of type $k$ occurred from source country $i$ to destination country $j$ during time step $t$, and $\bcalX_{t,ijk}=0$ otherwise. This binary representation allows us to apply our TSSDMN model to uncover latent temporal and cross-layer dynamics in international relations.
We focus on the 30 most prominent countries from 2009 to 2017. Treating countries as nodes and event types as layers, we construct a dynamic multilayer network with $n=30$, $T=95$ (monthly), and $K=4$. The tensor entry $\bcalX_{t,ijk}$ is set to 1 if an event of type $k$ occurred from country $i$ to $j$ during month $t$, and 0 otherwise. This binary formulation allows our model to uncover latent dynamics in international relations.

\begin{figure}[t]
    \centering
    \includegraphics[width=\columnwidth]{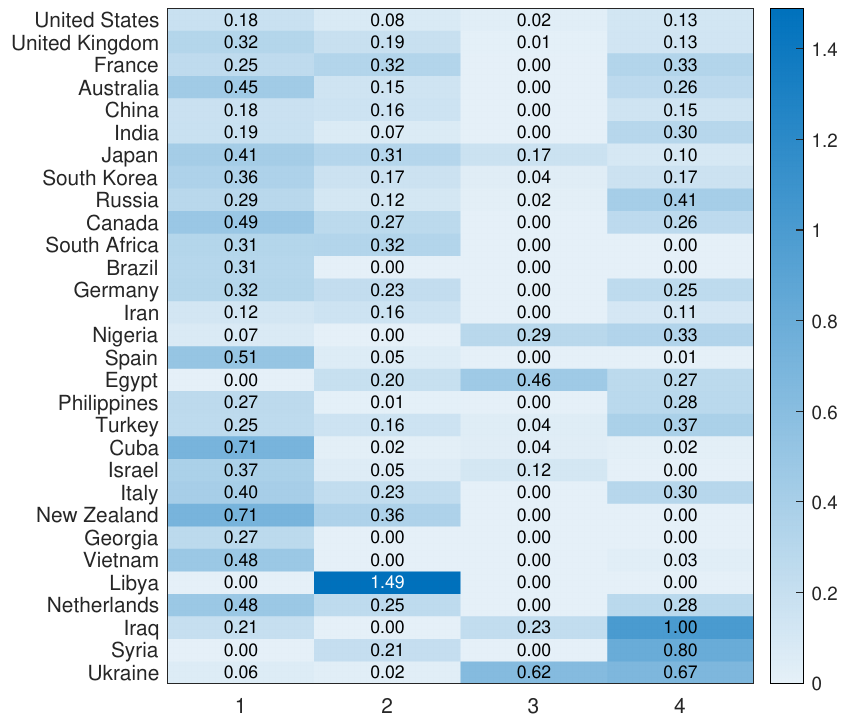}
    \caption{The estimated four latent factors of the ICEWS dataset.
    \label{fig:case1-community}}
\end{figure}

For the ICEWS dataset, we set the latent dimension $m=4$ for TSSDMN, based on the AIC-based model selection procedure described in Section~\ref{sec:parameter-selection}. The estimated latent factor matrix $\hat{\bC}_{1}$, which captures the behavior profiles of each country, is visualized in Figure~\ref{fig:case1-community}. Each column in $\hat{\bC}_{1}$ corresponds to a latent behavioral pattern, and each entry in row $i$, column $j$ reflects the degree to which country $i$ exhibits behavior pattern $j$. The latent factors are constrained to be nonnegative, allowing for straightforward interpretation of memberships. Factor 1 appears to capture general, baseline behavior patterns shared across many countries. Factor 2 highlights distinctive interaction patterns centered around Libya and its related countries, corresponding to the Libyan civil war and NATO intervention.
Factor 3 represents unique behaviors involving Egypt, Ukraine, and their neighboring countries, aligning with key events such as the Arab Spring and the 2013–2014 Ukrainian crisis.
Factor 4 is associated with Iraq, Syria, and Ukraine, and reflects patterns influenced by the Arab Spring and the subsequent regional instability.

\begin{figure}[t]
    \centering
    \includegraphics[width = \linewidth]{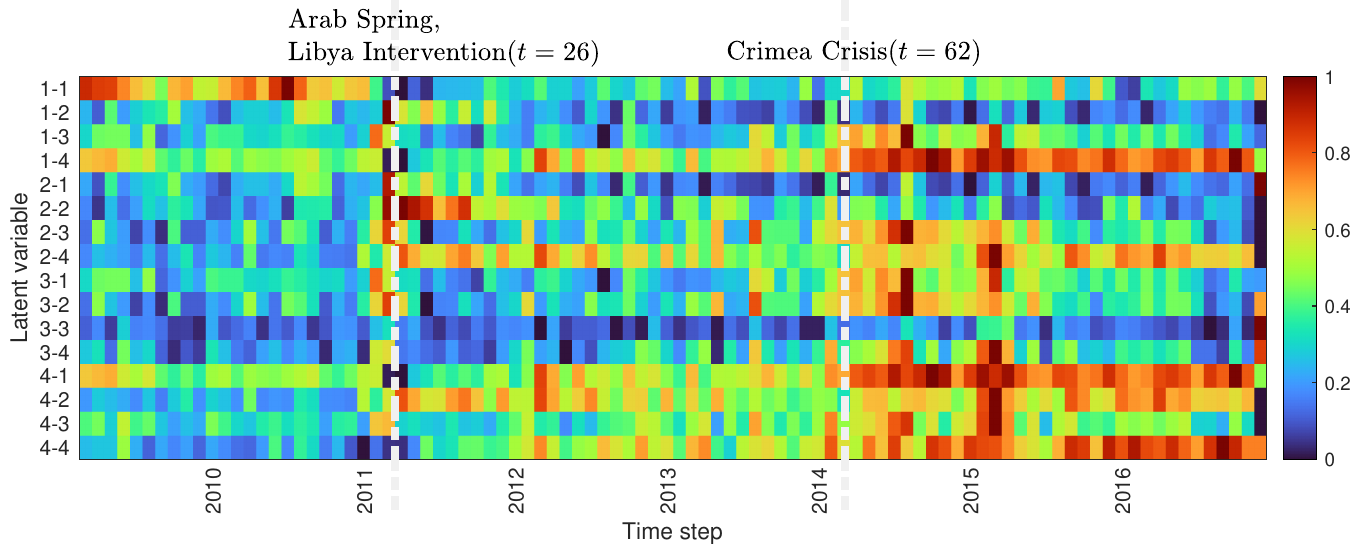}
    \caption{Dynamic interaction patterns between the latent factors for the ``verbal cooperation'' layer, scaled by each row of the ICEWS dataet. Row "a-b" represents the dynamic interactions from factor a to factor b.  %\chen{change time index to year 2009-2017 for all the figures, and change the y-axis label to ``1-1'', ``1-2'',...``4-4'' }
    \label{fig:verbal-cooperation}}
\end{figure}

Except for the first, general-purpose factor, the other three latent dimensions can be directly associated with major geopolitical events that occurred during the observed time period, demonstrating the model's capacity to uncover historically meaningful structure in international interactions. To further investigate the temporal dynamics of these latent factors, we visualize the interaction patterns among the latent factors over time within the "verbal cooperation" layer in Figure~\ref{fig:verbal-cooperation}. The interaction patterns exhibit a significant jump in March 2011 ($t=26$), corresponding to the onset of the military intervention in Libya and the escalation of the Arab Spring. Another notable jump is observed in February 2014 ($t=62$), coinciding with the Crimea Crisis in Ukraine. This shift reflects changes in communication dynamics, particularly between Ukraine and Russia.

 %The 5th row represents the low level of verbal cooperation between Libya and other countries. 
\begin{figure}[!t]
    \centering
    \includegraphics[width = \columnwidth]{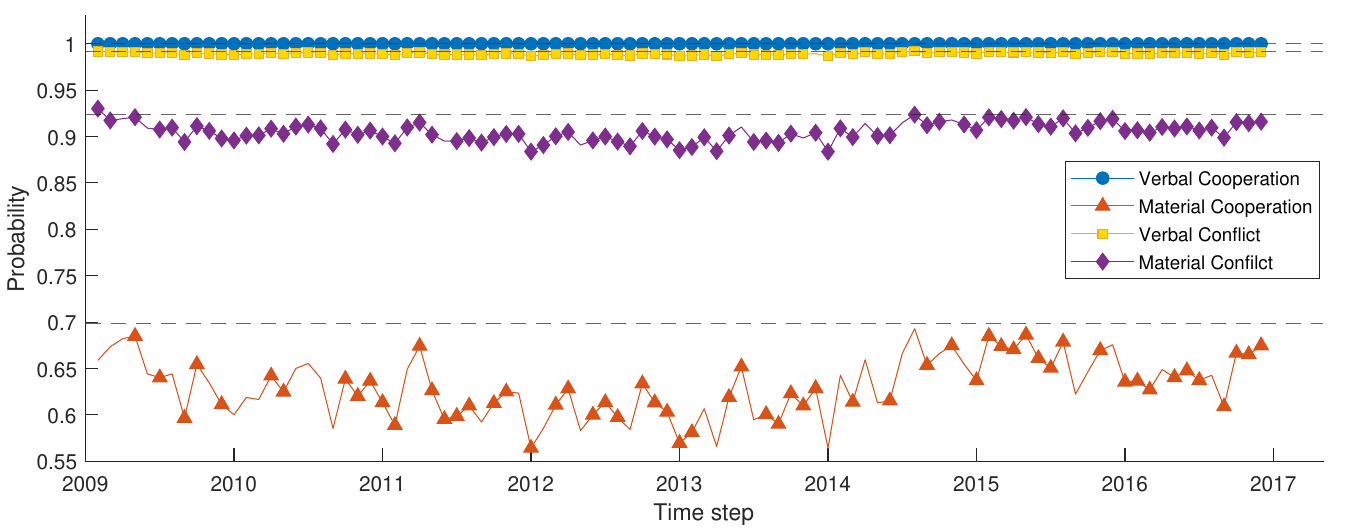}
    \caption{Connection probabilities between the United States ($i$) and Russia ($j$) of the ICEWS dataset. The dotted lines represent static probabilities calculated by $\bb_{ij}$ and the colored lines around the dotted lines are the estimated probabilities $p_{t,ijk}$.
    \label{fig:US-Russia-relation}}
\end{figure}

To illustrate the interpretability of our model at the country-pair level, we examine the dynamic and static relationships between two key actors: the United States and Russia. We begin by analyzing the static bias parameter $\bb_{ij} \in \bbR^4$, where $i$ and $j$ correspond to the United States and Russia. Its value is $[1, 0.66, 0.99, 0.91]$, 
capturing the baseline frequency of the four types of events between the two countries and indicating a consistently high frequency of verbal cooperation, along with substantial levels of both verbal and material conflict. This aligns with the complex diplomatic and adversarial history between the two nations. To explore temporal dynamics, Figure \ref{fig:US-Russia-relation} further plots the time-varying connection probabilities $p_{t,ijk}$ for all four event types between the United States and Russia over time. We see that the probabilities for verbal and material cooperation remain relatively stable throughout the time period. There is a slight decline in material cooperation observed between 2011 and 2014, corresponding to deteriorating relations during and after the Libya intervention. In 2014, material cooperation remains low, likely due to the Russia–Ukraine conflict and the annexation of Crimea. A modest rebound is visible in subsequent years, potentially attributable to diplomatic efforts such as the Minsk II agreement, which marked a partial de-escalation in the region.

Finally, we evaluate the predictive performance of all models by computing the AUC for time step $T+1$. For a fair comparison, we tune the hyperparameters of each baseline model to achieve their best possible predictive AUC on the ICEWS dataset. 
Table \ref{tab:ICEWS-result} summarizes the results across the four event-type layers. Our proposed model achieves the highest AUC in predicting verbal cooperation, verbal conflict, and material conflict. In the material cooperation layer, the AUC of TSSDMN is comparable to the best-performing AUC, showing only a marginal difference. This result further highlights the robustness and effectiveness of our model in capturing complex temporal and cross-layer dependencies in multilayer dynamic networks.

\begin{table}[h]
\centering
\begin{tabular}{ccccc}
%& \multicolumn{4}{c}{Preditive AUC} \\
\toprule
        Model    &  VCoo & MCoo & VCon & MCon\\
        \toprule
        \textbf{TSSDMN} & \textbf{0.847} & \underline{0.897} & \textbf{0.884} & \textbf{0.860} \\
        \textbf{MTR} & 0.746 & 0.794 & 0.787 & 0.771 \\
        \textbf{EDMN} & 0.797 & \textbf{0.902} & \underline{0.867} & \underline{0.857} \\
        \textbf{BDMN} & \underline{0.807} & 0.860 & 0.856 & 0.835 \\
        \textbf{DSBM} & 0.779 & 0.459 & 0.791 & 0.645 \\
\bottomrule
\end{tabular}
\caption{Prediction AUC of the ICEWS dataset, "VCoo", "MCoo", "VCon", and "MCon" refer to Verbal Cooperation, Material Cooperation, Verbal Conflict, and Material Conflict, respectively. 
\label{tab:ICEWS-result}}
\end{table}

\subsection{UNSW-NB15 Dataset}
\label{sec:case2}
The UNSW-NB15 dataset \cite{moustafa2015unsw} is a benchmark for network anomaly detection. We model IP addresses as nodes and communications as directed edges. Communications are categorized into four types (normal, fuzzers, exploits, reconnaissance), forming a four-layer dynamic network. This structure allows our model to uncover latent patterns and anomalous dynamics in cybersecurity data.
\begin{figure}[t]
    \centering
    \includegraphics[width=\linewidth]{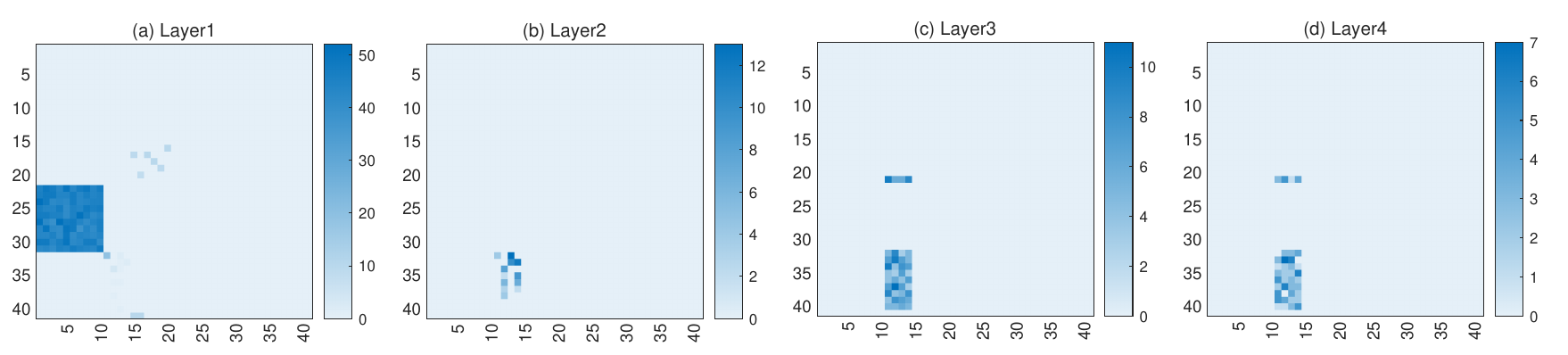}
    \caption{Accumulated frequencies of the four edge types over $T=60$ time steps of the UNSW-NB15 dataset.
    \label{fig:UNSW-NB15-frequency}}
\end{figure}

\begin{figure}[t]
    \centering
    \includegraphics[width=\linewidth]{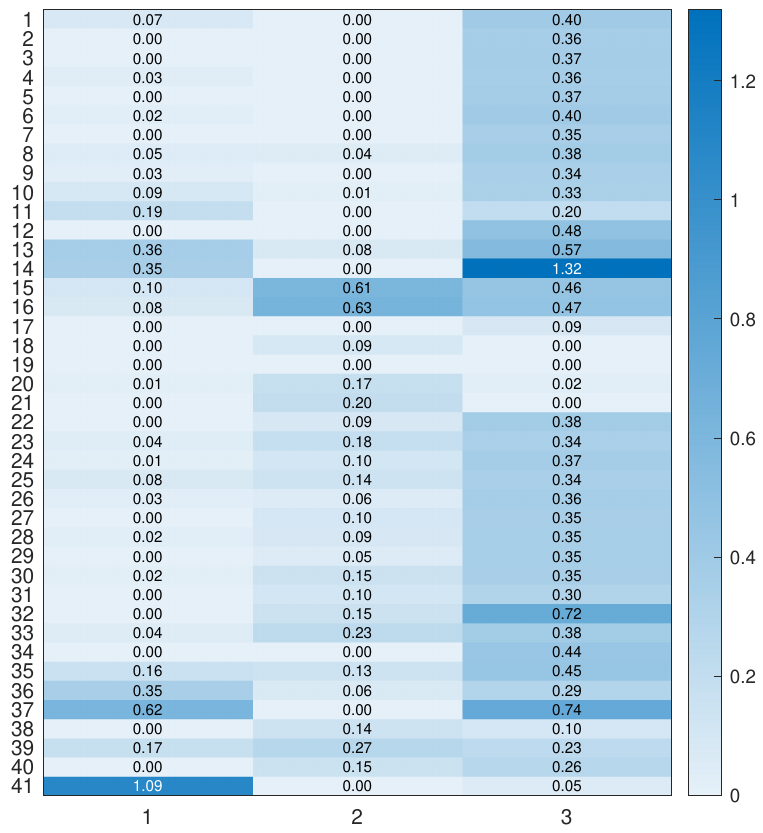}
    \caption{The estimated three latent factors of the UNSW-NB15 Dataset. \label{fig:IoT-community}}
\end{figure}
Our study focus on the 40 most active IP addresses and collects communication data at 10-second intervals over a total of 500 seconds. Within each 10-second window, a directed edge is created between two nodes if a communication occurs between the corresponding IP addresses. This results in the construction of a dynamic multilayer network with $n=40$ nodes, $K=4$ layers (communication types) and $T=60$ time steps. Layer 1 corresponds to normal communication and layers 2 to 4 correspond to attack types: fuzzers, exploits and reconnaissance, respectively. Figure~\ref{fig:UNSW-NB15-frequency} shows the accumulated frequency of connections across all four layers over the 60 time steps. In the normal communication layer, most traffic flows from nodes 22–32 to nodes 1–10, suggesting a typical client-server interaction pattern. In contrast, the three attack layers exhibit markedly different behavior. In particular, attack sources are concentrated among nodes 33–40, and their targets are primarily nodes 11–15, which may indicate attempted intrusions on a vulnerable subnet. Additionally, the overall frequency of attacks is significantly lower than that of normal communications, reflecting the sparse and intermittent nature of malicious activity.

\begin{figure}[t]
    \centering
    \includegraphics[width=\linewidth]{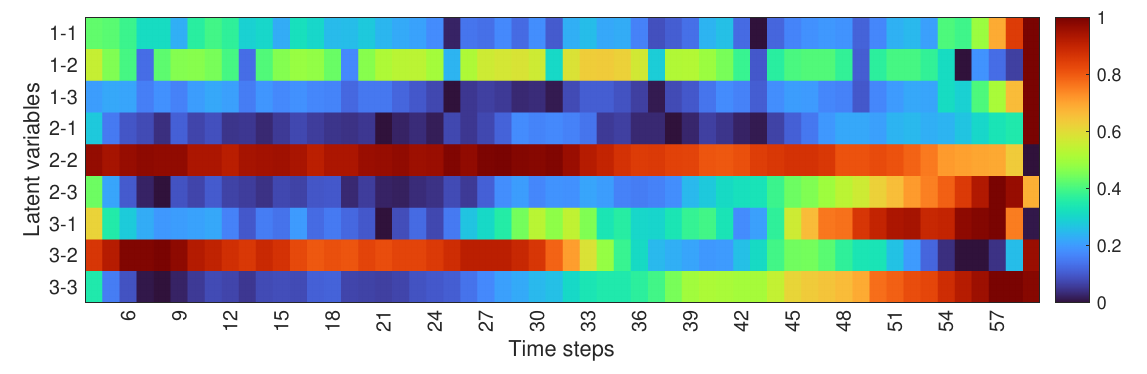}
    \caption{Dynamic interaction patterns between the latent factors for the ``Fuzzers'' layer, scaled by each row of the UNSW-NB15 dataset. \label{fig:IoT-latentvariables}}
\end{figure}

\begin{table}[t]
\centering
\begin{tabular}{ccccc}
%& \multicolumn{4}{c}{Predictive AUC} \\
\toprule
Model & Normal & Fuzzers & Exploits & Reconnaissance \\
\toprule
    \textbf{TSSDMN} & \textbf{0.995} & \textbf{0.999} & \textbf{0.992} & \textbf{0.991} \\
    \textbf{MTR} & 0.974 & 0.621 & 0.505 & 0.637 \\
    \textbf{EDMN} & 0.795 & 0.721 & 0.755 & 0.713 \\
    \textbf{BDMN} & \textbf{0.995} & \textbf{0.999} & \underline{0.990} & \underline{0.979} \\
    \textbf{DSBM} & 0.993 & 0.988 & 0.006 & 0.948 \\
\bottomrule
\end{tabular}
\caption{Prediction AUC of the UNSW-NB15 dataset
\label{tab:UNSW-NB15-result}}
\end{table}

For this dataset, we set the latent dimension $m=3$ for TSSDMN based on our AIC selection criteria. 
Figure \ref{fig:IoT-community} shows the three estimated latent factors $\bC_1$, representing three distinct latent behavioral patterns across the 40 nodes. Combined with the dynamic interaction patterns visualized in Figure~\ref{fig:IoT-latentvariables}, we can find several interesting temporal dynamics emerge among the latent factors. The interactions from Factor 1 to Factor 2 exhibit a clear periodicity with a cycle of approximately 6 time steps. Given the 10-second time resolution, this suggests that certain attack behaviors may recur once per minute, potentially indicating automated or scripted attack routines originating from nodes in Factor 1 and targeting those in Factor 2. The interactions from Factor 3 to Factor 2 are relatively strong during the early period ($t \leq 33$) but weaken significantly afterward ($t > 33$). This shift may reflect a temporal change in attack strategy or source activity, such as a wave of reconnaissance or exploit attempts concentrated in the earlier phase of the observation window.

Finally, we evaluate the predictive performance of all models on the UNSW-NB15 dataset by computing the AUC at time step $T+1$, following the same protocol as in Section~\ref{sec:case1}. The results are summarized in Table~\ref{tab:UNSW-NB15-result}.
Among all methods, TSSDMN consistently achieves the highest AUC, particularly excelling in the prediction of attack-related communication types. The second-best model is BDMN, which also demonstrates competitive accuracy across most layers. In contrast, the remaining two baselines—though they achieve reasonable AUC scores for predicting normal communication—perform poorly in predicting the three types of attacks. This discrepancy is likely due to the class imbalance between frequent normal communications and sparse attack events, which can adversely affect models that are not designed to account for such asymmetries.

\section{Conclusion} \label{sec:conclusion}
This paper introduces a novel tensor state space-based dynamic multilayer network model within the latent space model framework. It adopts a symmetric nonnegative Tucker decomposition to characterize the latent factors of nodes and inter-layer transitions, and integrates a tensor autoregressive structure to capture the temporal evolution of the network, accounting for both intra-layer and cross-layer dynamics. To enable scalable and efficient inference, we develop a variational EM algorithm, and propose an AIC-based criterion for automatic selection of the latent factor dimension. Theoretical analysis establishes the identifiability of the model, and the convergence of the variational EM.  Extensive simulation studies and two real-world case studies demonstrate the efficiency and superiority of our model compared with existing state-of-the-art methods.

Looking forward, several promising directions remain for future research. First, incorporating cross-layer edges into the model would allow for a more careful representation of interactions between different layers, which is essential for many real-world multilayer networks. Second, extending the model to handle weighted edges would enable the analysis of interaction strengths, enabling more nuanced inference in settings where edge intensity carries meaningful information.

%\bibliography{bibtex/bib/IEEEabrv.bib,bibtex/bib/IEEEexample.bib}{}
\bibliography{references}
\bibliographystyle{IEEEtran}
\newpage

\appendices

\section{Data generation details for 3 baselines}
\label{app:data-generation}
\begin{itemize}
    \item \textbf{EDMN:} Eigenmodel for dynamic multilayer networks proposed by \cite{loyal2023eigenmodel}.
    
    1. Denote number of nodes $n$, number of layers $K$ and variance $\sigma^2$. Setting number of factor variables $m=2$, variance of sociality effects $\tau = 0.1$;
    
    2. Generate $\lambda_{1h} = 2u_h - 1$, where $u_h \sim {\rm Bernoulli}(0.5)$;
    
    3. Generate the remaining homophily coefficients $\lambda_k \sim \mathcal{U}(-2,2)$;
    
    4. Generate initial state of sociality effects $\delta^{k,i}_1 \sim \mathcal{U}(-4,4)$. Generate $\delta^{k,i}_{t} \sim \mathcal{N}(\delta^{k,i}_{t-1}, \tau \mathcal{I}_k)$;
    
    5. Generate initial latent positions $\bz^i_1 \sim \mathcal{N}(0, 4 \mathcal{I}_m)$. Generate $\bz^i_t \sim \mathcal{N}(\bz^{i}_{t-1}, \sigma^2 \mathcal{I}_m)$;
    
    6. Centering the latent space $\tilde{\bz}^i_t = \bz^i_t - \frac{1}{n} \sum_{j=1}^n \bz^j_t$;
    
    7. Generate observation $\bcalX_{t,ijk}$ by ${\rm logit}(P(\bcalX_{t,ijk} = 1)) = \delta^{i,k}_{t} + \delta^{j, k}_{t} + (\tilde{\bz}^i_t)^T \Lambda_k \tilde{\bz}^j_t$.
    
    \item \textbf{BDMN:} Bayesian dynamic multilayer network proposed by \cite{durante2017bayesian}.
      
    1. Give number of nodes $n$, number of layers $K$, variance $\sigma$. Setting number of factor variables $m=2$;
    
    2. Generate the latent variable $\Gamma^t_{i,j,k}$;
        
    3. Generate observations $Y_{i,j,k}^t$ by $Y_{i,j,k}^t\sim Bern(\Gamma_{i,j,k}^t)$.
    % \item \textbf{DSBM:} Dynamic stochastic block model proposed by \cite{xu2014dynamic}.
    
    % 1. Give number of nodes $n$, number of layers $K$, variance $\sigma$. Setting the number of classes to be $\frac{n}{5}$, the probability of classes changed in each time step $p$ to be 0.1;
    
    % 2. Setting the variance of all states to be $\sigma ^ 2$ and the covariance of between any state and any neighboring state to be $\frac{\sigma ^ 2}{4}$, Generate covariance matrix $\Sigma$;
    
    % 3. Generate initial state $\psi^0$, classes $c^0_i$. For any $t\geq1$, $\psi^t \sim \mathcal{N}(\psi^{t-1}, \Sigma)$. Generate $c^{t}$ from $c^{t-1}$ and change classes of $c^{t-1}$ with probability $p$;
    
    % 4. Generate $\gamma^t_{i,j} = logit(\frac{\psi^t_{c^t_i,c^t_j}}{1 - \psi^t_{c^t_i,c^t_j}})$. Generate observation $y^t_i \sim Bern(\gamma_t^i)$. 
\end{itemize}

\section{Proof of Theorems}
\subsection{Proof of Theorem \ref{the:identifiability}}
\label{appx:proof-identifiability}

Under Assumption \ref{ass:stationarity}, the tensor autoregressive process admits a stationary distribution due to the stability condition $\rho(\bA_i) < 1$. This guarantees that the series $\{\bcalZ_t\}$ converges to a zero-mean Gaussian process as $t \to \infty$, with the covariance structure satisfying the Lyapunov equation:
\begin{equation*}
    \text{vec}(\bSigma_\infty) = (\bA_3 \otimes \bA_2 \otimes \bA_1)\text{vec}(\bSigma_\infty) + \sigma^2\text{vec}(\bI),
\end{equation*}
where $\bSigma_\infty = \mathbb{E}[\text{vec}(\bcalZ_t)\text{vec}(\bcalZ_t)^T]$ denotes the stationary covariance matrix. For observationally equivalent parameters $\Theta$ and $\Theta'$, the log-odds tensor must satisfy $\bGamma_t = \bGamma'_t$ for all $t$. Substituting the Tucker decomposition from Equation (\ref{equ:beta}) yields:
\begin{equation*}
    \bcalB + \bcalZ_t \times_1 \bC_1 \times_2 \bC_1 \times_3 \bC_2 = \bcalB' + \bcalZ'_t \times_1 \bC'_1 \times_2 \bC'_1 \times_3 \bC'_2.
\end{equation*}
As $t \to \infty$, the dynamic components converge to their stationary distributions, with $\lim_{t\to\infty} \mathbb{E}[\bcalZ_t] = \mathbf{0}$ and $\lim_{t\to\infty} \mathbb{E}[\bcalZ'_t] = \mathbf{0}$. Taking expectations on both sides of the log-odds equation gives:
\begin{equation*}
    \bcalB = \bcalB' + \lim_{t\to\infty} \mathbb{E}[(\bcalZ'_t \times_1 \bC'_1 \times_2 \bC'_1 \times_3 \bC'_2) - (\bcalZ_t \times_1 \bC_1 \times_2 \bC_1 \times_3 \bC_2)].
\end{equation*}
The vanishing expectation of dynamic terms leaves $\bcalB = \bcalB'$. 

From the log-odds equality $\bGamma_t = \bGamma'_t$ and the stationary limit $\bcalB = \bcalB'$, we obtain the core relationship:
\[
\bcalZ_t \times_1 \bC_1 \times_2 \bC_1 \times_3 \bC_2 = \bcalZ'_t \times_1 \bC'_1 \times_2 \bC'_1 \times_3 \bC'_2.
\]
Under Assumption \ref{ass:pure-source}, the nonnegative loading matrix $\bC_1$ satisfies the permuted anchored condition $\bC_1 = \bPi_1[\bI, \mathbf{U}^T]^T\bPi_2\bD_0$. By Proposition 6 of \cite{zhou2015efficient}, this structure guarantees uniqueness up to permutation and scaling:
\[
\bC_1 = \bC'_1 \bPi \bD_1,\quad \bPi \in \mathcal{P}^{m\times m}, \ \bD_1 \in \mathcal{D}^{m\times m}_+.
\]
The Frobenius norm condition $\|\bC_1\|_F = \|\bC'_1\|_F$ enforces $\bD_1 = \bI$, leaving only permutation ambiguity $\bPi$. For the layer transition matrix $\bC_2$, the relationship:
\[
\bC_2 = \bC'_2 \bR,\quad \bR^T\bR = \bI
\]
emerges from the covariant structure of $\bcalZ_t$ under orthogonal transformations. The equivalence:
\[
\bcalZ_t = \bcalZ'_t \times_1 \bPi \times_2 \bPi \times_3 \bR
\]
preserves the interaction patterns while allowing latent dimension permutation ($\bPi$) and layer-wise rotations ($\bR$).  
\qed
\endproof

\subsection{Minimax Lower Bound Analysis(Proof of Theorem 2)}
\label{appx:minimax}

\subsubsection{Hypothesis Construction for the Core Tensor}

Our goal is to construct a set of hypotheses for the core tensor sequence $\{\boldsymbol{\mathcal{Z}}_t\}$ that are hard to distinguish from a null hypothesis (e.g., $\boldsymbol{\mathcal{Z}}_t = \mathbf{0}$ for all $t$). We adapt the logic from tensor estimation by constructing sparse perturbations directly on the core tensor.

Let the dimensionality of the core tensor be $m^2K$. We consider the vector space $\mathbb{R}^{m^2K}$ by vectorizing the tensors. We apply Lemma 1 to the binary space $\{0, 1\}^{m^2K}$.

\begin{lemma}[Massart, 2007, Lemma 4.10]
Let $\Omega = \{0, 1\}^{m^2K}$ and $1 \le s \le m^2K/4$. There exists a subset $\{\mathbf{w}^{(1)}, \dots, \mathbf{w}^{(M)}\} \subset \Omega$ such that:
\begin{enumerate}
  \item  $\|\mathbf{w}^{(l)}\|_0 = s$ for all $1 \le l \le M$.
  \item  $\|\mathbf{w}^{(l)} - \mathbf{w}^{(j)}\|_0 \ge s/2$ for all $0 \le l \neq j \le M$.
  \item  $\log M \ge c s \log(m^2K/s)$ for a constant $c \ge 0.233$.
\end{enumerate}
\end{lemma}

For each binary vector $\mathbf{w}^{(l)} \in \{\mathbf{w}^{(1)}, \dots, \mathbf{w}^{(M)}\}$, we construct a base perturbation tensor $\mathcal{W}^{(l)} \in \mathbb{R}^{m \times m \times K}$ by reshaping $\mathbf{w}^{(l)}$. The elements of $\mathcal{W}^{(l)}$ are either 0 or 1. The set of hypotheses for the core tensor at a single time point is then defined as:
\begin{equation*}
\Theta_{\boldsymbol{\mathcal{Z}}} = \left\{ \boldsymbol{\mathcal{Z}}^{(l)} = \epsilon \mathcal{W}^{(l)} \mid l=1, \dots, M \right\}
\end{equation*}
where $\epsilon$ is a small perturbation magnitude to be determined later. The null hypothesis is $\boldsymbol{\mathcal{Z}}^{(0)} = \mathbf{0}$.

The crucial distance for our problem is not on $\boldsymbol{\mathcal{Z}}_t$ directly, but on the resulting log-odds tensor $\boldsymbol{\Gamma}_t = \boldsymbol{\mathcal{Z}}_t \times_1 \mathbf{C}_1 \times_2 \mathbf{C}_1 \times_3 \mathbf{C}_2$. The distance between two hypotheses, indexed by $l$ and $j$, is:
\begin{align*}
d_0^2(\boldsymbol{\mathcal{Z}}^{(l)}, \boldsymbol{\mathcal{Z}}^{(j)}) &= \|\boldsymbol{\Gamma}^{(l)} - \boldsymbol{\Gamma}^{(j)}\|_F^2 \\
&= \epsilon^2 \| (\mathcal{W}^{(l)} - \mathcal{W}^{(j)}) \times_1 \mathbf{C}_1 \times_2 \mathbf{C}_1 \times_3 \mathbf{C}_2 \|_F^2
\end{align*}

To rigorously bound this distance, we analyze the Frobenius norm using the vectorization of the tensor product. Let $\Delta\mathcal{W} = \mathcal{W}^{(l)} - \mathcal{W}^{(j)}$. The log-odds distance can be expressed as:
\begin{align*}
\|\boldsymbol{\Gamma}^{(l)} - \boldsymbol{\Gamma}^{(j)}\|_F^2 &= \epsilon^2 \| \text{vec}(\Delta\mathcal{W} \times_1 \mathbf{C}_1 \times_2 \mathbf{C}_1 \times_3 \mathbf{C}_2) \|_2^2 \\
&= \epsilon^2 \| (\mathbf{C}_2 \otimes \mathbf{C}_1 \otimes \mathbf{C}_1) \text{vec}(\Delta\mathcal{W}) \|_2^2
\end{align*}
where $\otimes$ denotes the Kronecker product.

Under Assumption 4, we can bound the norm from above and below using the singular values of the Kronecker product matrix $\mathbf{C} = \mathbf{C}_2 \otimes \mathbf{C}_1 \otimes \mathbf{C}_1$.
The largest and smallest singular values of $\mathbf{C}$ are $\sigma_{\max}(\mathbf{C}) = \sigma_{\max}(\mathbf{C}_2)\sigma_{\max}(\mathbf{C}_1)^2 \le \lambda_{2\max}\lambda_{1\max}^2$ and $\sigma_{\min}(\mathbf{C}) = \sigma_{\min}(\mathbf{C}_2)\sigma_{\min}(\mathbf{C}_1)^2 \ge \lambda_{2\min}\lambda_{1\min}^2$.

This provides the necessary bounds. For the KL-divergence condition (comparing hypothesis $l$ to the null hypothesis $0$), we need an upper bound:
\begin{align*}
d_0^2(\boldsymbol{\mathcal{Z}}^{(l)}, \boldsymbol{\mathcal{Z}}^{(0)}) &\le \epsilon^2 \sigma_{\max}(\mathbf{C})^2 \|\text{vec}(\mathcal{W}^{(l)})\|_2^2 \\
&\le \epsilon^2 (\lambda_{2\max}\lambda_{1\max}^2)^2 \|\mathbf{w}^{(l)}\|_0 \\ 
&= (\lambda_{2\max}\lambda_{1\max}^2)^2 \epsilon^2 s
\end{align*}
For the hypothesis separation condition ($l \neq j$), we need a lower bound:
\begin{align*}
d_0^2(\boldsymbol{\mathcal{Z}}^{(l)}, \boldsymbol{\mathcal{Z}}^{(j)}) &\ge \epsilon^2 \sigma_{\min}(\mathbf{C})^2 \|\text{vec}(\Delta\mathcal{W})\|_2^2 \\
&\ge \epsilon^2 (\lambda_{2\min}\lambda_{1\min}^2)^2 \|\mathbf{w}^{(l)} - \mathbf{w}^{(j)}\|_0 \\&\ge \frac{(\lambda_{2\min}\lambda_{1\min}^2)^2}{2} \epsilon^2 s
\end{align*}

\subsubsection{Hypothesis Construction for Total Variation Denoising}

To model temporal dynamics, we construct hypotheses that vary over time. We partition the time interval $\{1, \dots, T\}$ into $m_t$ blocks $S_1, \dots, S_{m_t}$, each of size $k_t$, such that $T \approx m_t k_t$.

We use another application of Lemma 1 to define which time blocks are perturbed. Let $\Omega_{m_t} = \{\phi^{(1)}, \dots, \phi^{(M_0)}\} \subset \{0, 1\}^{m_t}$ be a set of binary vectors where each $\phi$ has Hamming weight $\|\phi\|_0 = s_0 $ with $ s_0 \leq m_t/4$ and the Hamming distance between any two distinct vectors is at least $s_0/2$. The number of such vectors is $M_0 \ge \exp(c s_0 \log(m_t/s_0))$.

The full hypothesis set $\Theta$ for the sequence $\{\boldsymbol{\mathcal{Z}}_t\}$ is constructed by combining the spatial perturbations $\mathcal{W}^{(l)}$ with the temporal perturbations $\phi$:
\begin{equation*}
\begin{split}
\Theta = \Bigg\{ 
\boldsymbol{\mathcal{Z}}^{(\mathbf{l}, \phi)} : 
\boldsymbol{\mathcal{Z}}_t = 
\begin{cases} 
\epsilon \mathcal{W}^{(l_j)} & \text{if } t \in S_j \text{ and } \phi_j=1 \\
\mathbf{0} & \text{if } t \in S_j \text{ and } \phi_j=0
\end{cases}
, \\ \text{ for } \mathbf{l}=(l_1, \dots, l_{s_0}) \in \{1,\dots,M\}^{s_0}, \phi \in \Omega_{m_t}
\Bigg\}
\end{split}
\end{equation*}
Here, for each activated block $S_j$ (where $\phi_j=1$), we assign a perturbation tensor $\mathcal{W}^{(l_j)}$ from our base set. The null hypothesis $\boldsymbol{\mathcal{Z}}^{(0)}$ corresponds to $\boldsymbol{\mathcal{Z}}_t = \mathbf{0}$ for all $t$.

The size of this hypothesis set is $|\Theta| = M_0 M^{s_0}$, and for $\mathcal{U},\mathcal{V} \in \Theta$, we have:
\begin{equation*}
D_{KL}(P_{\mathcal{U}},P_{\mathcal{V}}) \leq d(\mathcal{U}, \mathcal{V})= \sum_{t=1}^T d_0^2(\mathcal{U},\mathcal{V}) =  \epsilon^2 s_0 s k_t \lambda_{\max}^3  
\end{equation*}

To derive the minimax lower bound, we apply Fano's inequality, presented here as Lemma 2.

\begin{lemma}[Theorem 2.5 in Tsybakov, 2008]
Suppose $M_{tot} \ge 2$ and $(\Theta, d)$ contains elements $\{\theta_0, \dots, \theta_{M_{tot}}\}$ such that for any $0 \le i \neq j \le M_{tot}$, $d(\theta_i, \theta_j) \ge 2\delta > 0$, and furthermore $\frac{1}{M_{tot}}\sum_{i=1}^{M_{tot}} D_{KL}(P_{\theta_i} \| P_{\theta_0}) \le \alpha \log M_{tot}$ with $0 < \alpha < 1/8$. Then we have:
\begin{equation*}
\inf_{\hat{\theta}} \sup_{\theta \in \Theta} P(d(\hat{\theta}, \theta) \ge \delta) \ge \frac{\sqrt{M_{tot}}}{1+\sqrt{M_{tot}}} \left(1-2\alpha-\sqrt{\frac{2\alpha}{\log M_{tot}}}\right)
\end{equation*}
\end{lemma}

The KL-divergence condition from Lemma 2 requires $\frac{1}{M_{tot}}\sum D_{KL}(P_{\theta_i} \| P_{\theta_0}) \le \alpha \log M_{tot}$. Plugging in our bound:
\begin{equation*}
\frac{1}{2} \epsilon^2 s_0 k_t \lambda_{\max}^3 s \le \alpha \log(M_0 M^{s_0}) = \alpha(\log M_0 + s_0 \log M)
\end{equation*}
Using Lemma 1, $\log M \ge c s \log(m^2K/s)$ and $\log M_0 \ge c s_0 \log(m_t/s_0)$. To simplify, we require the condition to hold, which gives us an upper bound on how large $\epsilon$ can be. We choose $\epsilon$ to be at the limit of this condition to maximize the separation distance $\delta$:
\begin{equation*}
\epsilon^2 \lambda_{1 \max}^4 \lambda_{2 \max}^2 s s_0 k_t \leq 2\alpha [cs s_0\log (m^2K/s) + c s_0 \log (m_t/s_0)],
\end{equation*}
taking $\alpha < 1/8, s_0=m_t/4,s=m^2K/4$ 
\begin{equation*}
\lambda_{1 \max}^4 \lambda_{2 \max}^2 T s \epsilon^2 \leq \frac{(m^2K+4) m_t \log 4}{16}
\end{equation*}
where $c=0.233$.

\subsubsection{Minimax rate for TSD smoothness}
Based on our construction, since $\|\mathbf{A}_1\|_2,\|\mathbf{A}_2\|,\|\mathbf{A}_3\|_2\leq 1$ we need $2(m_t-1)s\epsilon < L$. This condition must hold for our constructed hypotheses to be valid members of the parameter space $\text{TDS}(\mathcal{L})$. We now consider different cases based on the interplay between the smoothness budget $L$ and other model parameters.

The constraint on $\epsilon$ can be written as:
\begin{equation*}
\epsilon < \frac{L}{2(m_t-1)s}
\end{equation*}
This inequality imposes an upper bound on the perturbation magnitude $\epsilon$ to ensure that our constructed hypotheses belong to the temporal smoothness class $\text{TDS}(\mathcal{L})$.

By substituting the upper bound on $\epsilon$ into the KL-divergence condition derived previously, we obtain a relationship between the temporal smoothness budget $L$ and the number of temporal blocks $m_t$. This inequality delineates the feasible region for our hypothesis testing framework:
\begin{equation*}
\frac{L^2 \lambda_{1 \max}^4 \lambda_{2 \max}^2 T }{4(m_t-1)^2s} \leq \frac{(m^2K + 4) m_t \log 4}{16}
\end{equation*}

Rearranging this inequality provides a condition that determines whether the temporal variation $L$ is large enough to dominate the estimation error, which guides our case analysis.
\begin{equation*}
\begin{split}
& 4m_t(m_t-1)^2 (m^2K / 4)(m^2K+4) \log 4 \\
& \qquad > 16 L^2 T \lambda^4_{1 \max} \lambda^2_{2 \max}.
\end{split}
\end{equation*}

\textbf{Case 1:} This corresponds to the regime where the temporal variation is large, allowing for frequent changes. We select the number of temporal blocks $m_t=T$ to capture the finest possible temporal resolution.
If $L>c_1 (T-1) m^2 K \lambda^{-2}_{1 \max} \lambda^{-1}_{2 \max} $ that satisfies $4T(T-1)^2 (m^2K / 4)(m^2K+4) \log 4 \leq 16 L^2 T \lambda^4_{1 \max} \lambda^2_{2 \max}$, then we choose $m_t=T$ and $\epsilon=c' \lambda^{-2}_{1 \max} \lambda^{-1}_{2 \max}$ satisfying $2(m_t-1)s\epsilon < L$. In this case, the separation distance for the hypotheses leads to a lower bound on the risk, which is given by:
\begin{equation*}
\frac{1}{2}\lambda_{1,\min}^4 \lambda_{2,\min}^2  s s_0 k_t \epsilon^2 \gtrsim m^2 K T
\end{equation*}

\textbf{Case 2:} This represents the intermediate regime where the smoothness constraint $L$ is neither too large nor too small.
If $c_0 T^{-1/2} m^2 K \lambda^{-2}_{1 \max} \lambda^{-1}_{2 \max}< L < c_1 (T-1) m^2 K \lambda^{-2}_{1 \max} \lambda^{-1}_{2 \max}$ satisfying that  
\begin{equation*}
\frac{L^2 \lambda_{1 \max}^4 \lambda_{2 \max}^2 T }{4 \cdot 3^2 s} \leq \frac{(m^2K + 4) 4 \log 4}{16}
\end{equation*}
, we choose an optimal number of temporal blocks $m_t \leq L^{2/3} T^{1/3} m^{-4/3} \lambda_{1,\max}^{4/3} \lambda_{2\max}^{2/3} K^{-2/3}$ that balances the trade-off between temporal resolution and statistical distinguishability.

In this case, the magnitude of the perturbation is set to:
\begin{equation*}
\epsilon = L^{1/3} T^{-1/3} m^{-2/3} K^{-1/3} \lambda_{1,\max}^{-4/3} \lambda_{2 \max}^{-2/3}
\end{equation*}
This choice of $\epsilon$ and $m_t$ results in the following lower bound on the estimation error, which depends on the smoothness budget $L$.
\begin{equation*}
\frac{1}{2}\lambda_{1,\min}^4 \lambda_{2,\min}^2  s s_0 k_t \epsilon^2 \gtrsim L^{2/3} T^{1/3} m^{2/3} K^{1/3} \lambda_{1,\min}^{4/3} \lambda_{2,\min}^{2/3}
\end{equation*}

\textbf{Case 3:} This is the high smoothness regime, where $L$ is small, indicating that the core tensor sequence evolves very slowly.
If $L \leq c_0 T^{-1/2} m^2 K \lambda^{-2}_{1 \max} \lambda^{-1}_{2 \max}$, we set $m_t=1$, which treats the entire time series as a single block. The perturbation magnitude is chosen as $\epsilon=T^{-1/2}\lambda^{-2}_{1 \max} \lambda^{-1}_{2 \max}$. In this setting, the minimax risk is determined by:
\begin{equation*}
\frac{1}{2}\lambda_{1,\min}^4 \lambda_{2,\min}^2  s s_0 k_t \epsilon^2 \gtrsim m^2K
\end{equation*}
By combining the lower bounds from these three distinct regimes, we establish the overall minimax rate for estimating the log-odds tensor $\mathbf{\Gamma}_t$.
Combine these three cases, we have
\begin{equation*}
\begin{split}
& \inf_{\{\hat{\boldsymbol{\mathcal{Z}}}_t\}} \sup_{\{\boldsymbol{\mathcal{Z}}_t\} \in \text{TDS}(\mathcal{L})} \mathbb{E} \left[ \frac{1}{Tn^2K} \sum_{t=1}^T \|\mathbf{\Gamma}_t - \hat{\mathbf{\Gamma}}_t\|_F^2 \right] \\
& \qquad \gtrsim \min \left\{\frac{m^2}{n^2T}, \frac{L^{2/3}m^{2/3}}{T^{2/3}n^{4/3}K^{1/3}}, \frac{m^2}{n^2}\right\}
\end{split}
\end{equation*}

\subsection{Proof of Theorem \ref{the:convergence}}
\label{appx:proof-convergence}

We establish the posterior convergence rate by applying the general theory of posterior contraction for fractional posterior distributions. 

Assume the data is generated according to model (1) with the true latent trajectory $\bZ^{*}=[\bz_1^{*}, \ldots, \bz_T^{*}]\in \bbR^{m^2K \times T}$. We define the $\epsilon$-neighborhood for KL divergence centered at $\bZ^{*}$ as:
\begin{equation*}
\begin{array}{r}
B_{n, T}\left(\bZ^* ; \epsilon\right):=\left\{\bZ: \int p_{\bZ^*} \log \left(\frac{p_{\bZ^*}}{p_{\bZ}}\right) d \mu \leq n^2K T \epsilon^2\right. \\
\left.\int p_{\bZ^*} \log ^2\left(\frac{p_{\bZ^*}}{p_{\bZ}}\right) d \mu \leq n^2K T \epsilon^2\right\}
\end{array}
\end{equation*}
where $\mu$ denotes the Lebesgue measure. 

The following lemma establishes the foundation for our posterior contraction analysis:

\begin{lemma}[Posterior Contraction, Bhattacharya et al., 2019]
\label{lemma:postcontraction}
Fix $\alpha\in (0,1)$. Suppose $\epsilon$ satisfies $n^2\epsilon^2\geq 2$ and $\Pi\left(B_{n, T}\left(\bZ^*, \epsilon\right)\right) \geq e^{-n^2K T \epsilon^2}$. Then for any $D\geq 2$ and $t>0$,
\begin{equation*}
\Pi_\alpha\left(\left.\frac{1}{n^2K T} D_\alpha\left(\bZ, \bZ^*\right) \geq \frac{D+3 t}{1-\alpha} \epsilon^2 \right\rvert\, \mathcal{X}\right) \leq e^{-t n^2K T \epsilon^2}
\end{equation*}
holds with probability at least $1-2 /\left\{(D-1+t)^2 n^2K T \epsilon^2\right\}$.
\end{lemma}

%\textbf{Step 1: Establishing the Set Inclusion}

To apply Lemma \ref{lemma:postcontraction}, we need to bound the prior probability of the neighborhood $B_{n, T}\left(\bZ^* ; \epsilon\right)$. For the Bernoulli likelihood, by Lemma 14 in \cite{zhao2024structured}, we have:
\begin{equation*}
\max\{D_{KL}(p_{\bZ}, p_{\bZ^{*}}),V_{2}(p_{\bZ}, p_{\bZ^{*}})\}\leq\sum_{t=1}^{T}\sum_{i=1}^{n^2K}(\bC_{i}\bz_{t}-\bC_{i}\bz_{t}^{*})^2
\end{equation*}

Therefore, it suffices to lower bound the prior probability of the set:
\begin{equation*}
\left\{\sum_{t=1}^{T}\sum_{i=1}^{n^2K}(\bC_{i}\bz_{t}-\bC_{i}\bz_{t}^{*})^2\leq n^2KT\epsilon^2\right\}
\end{equation*}

Let $\tilde{\bC}_{ijk,:}=\bC_{(i-1)nK+(j-1)K+k,:}$ denote the row of the Kronecker product matrix. By establishing a sequence of set inclusions and assuming $\max_{ijk}\|\tilde{\bC}_{ijk,:}\|_{2}\leq C_{1}$, we obtain:
\begin{equation*}
\left\{\max_{t}\|\bz_{t}-\bz_{t}^{*}\|_{2}^{2}\leq \frac{\epsilon^2}{C_{1}^2}\right\} \subset \left\{\sum_{t=1}^{T}\sum_{i=1}^{n^2K}(\bC_{i}\bz_{t}-\bC_{i}\bz_{t}^{*})^2\leq n^2KT\epsilon^2\right\}
\end{equation*}

%\textbf{Step 2: Decomposition Using Model Structure}

Let $\epsilon_{0}=\frac{\epsilon}{C_{1}}$ and define the following events:
\begin{align}
E_{0} &= \left\{\max_{t}\|\bz_{t}-\bz_{t}^{*}\|_{2}\leq \epsilon_{0}\right\}, \\
E_{1} &= \left\{\max_{t\geq 2}\|(\bz_{t}-\bA^{t-1}\bz_{1})-(\bz_{t}^{*}-\bA^{t-1}\bz_{1}^{*})\|_{2}\leq \epsilon_{0}\right\}, \\
E_{2} &= \left\{\|\bz_{1}-\bz_{1}^{*}\|_{2}\leq \epsilon_{0}\right\}
\end{align}

By the independence structure of our model, we have:
\begin{equation*}
\Pi(E_0)\geq\Pi(E_{1})\Pi(E_{2})
\end{equation*}
where $\tilde{\bz}_{t}=\bz_{t}-\bA^{t-1}\bz_{1}$ represents the detrended process.

%\textbf{Step 3: Bounding $\Pi(E_1)$ via Gaussian Process Theory}

For the temporal dynamics component, note that:
\begin{equation*}
\tilde{\bz}_{t}=\bA^{t-2}\boldsymbol{\varepsilon}_{2}+\cdots+\bA\boldsymbol{\varepsilon}_{t-1}+\boldsymbol{\varepsilon}_{t}\sim \calN(\mathbf{0},\bSigma_{1})
\end{equation*}

Applying multivariate Gaussian concentration inequalities through Anderson's lemma:
\begin{align}
\Pi(E_{1}) &\geq \exp\left(-\frac{(\bz^{*})^T\bSigma_{1}^{-1}\bz^{*}}{2}\right)\Pi\left(\sup_{t\geq 2}\|\tilde{\bz}_{t}\|_{2}\leq \epsilon_{0}\right)
\end{align}

Since $\bz_{t}^{*}=\bA\bz_{t-1}^{*}+\varepsilon_{t}$ with $\varepsilon_{t}\sim \calN(\mathbf{0}, \sigma^2\bI_{m^2K})$, we have:
\begin{equation*}
-\frac{(\bz^{*})^T\bSigma_{1}^{-1}\bz^{*}}{2}=-\sum_{t=1}^{T}\frac{\|\bz_{t}^{*}-\bA\bz_{t-1}^{*}\|_{2}^2}{2\sigma^2}
\end{equation*}

To bound the supremum probability, we construct a Gaussian process $\{\tilde{z}(s)\}_{s \in [0,1]}$ through linear interpolation of $(\tilde{\bz}_{2},\ldots,\tilde{\bz}_{T})$. For each dimension $i$, the variance function satisfies $\sigma_{i}^2(h)=E[(z_{i}(s+h)-z_{i}(s))^2]=hT\sigma^2$, which is concave in $h$. The cross-covariance is bounded by $\sup_{s\in [0,1]}|\cov(z_{i}(s),z_{j}(s))|\leq \rho T\sigma^2$ for $i \neq j$.

By Lemma 13 in \cite{zhao2024structured}, we obtain:
\begin{equation*}
P\left(\sup_{0\leq s\leq 1}\|\bz(s)\|_{2}\leq \epsilon_0\right)\geq C_{2}\exp\left(-C_3\frac{m^2KT\sigma^2}{\epsilon_0^2(1-\rho)}\right)
\end{equation*}

Under the temporal smoothness condition $\|\bz_{t}^{*}-\bA\bz_{t-1}^{*}\|_{2}\leq \frac{C_0 L}{T}$ for all $t=2,\ldots,T$ with $L=o(m^2KT)$, we get:
\begin{equation*}
\Pi(E_{1})\geq  C_2\exp\left[-\frac{C_0^2L^2}{2T\sigma^2}-C_3\frac{m^2KT\sigma^2}{\epsilon_0^2}\right]
\end{equation*}

Optimally choosing $\sigma^2=\frac{\epsilon_0 L}{m\sqrt{K}T}$ yields:
\begin{equation*}
\log(\Pi(E_{1}))\geq -\frac{Lm\sqrt{K}}{\epsilon_0}
\end{equation*}

%\textbf{Step 4: Bounding $\Pi(E_2)$ for Initial Condition}

For the initial state concentration, using the Gaussian prior $\bz_{1} \sim \calN(\mathbf{0}, \omega^2\bI_{m^2K})$:
\begin{align}
\Pi(E_{2}) &= \Pi(\|\bz_{1}-\bz_{1}^{*}\|_{2}\leq \epsilon_{0}) \\
&\geq \exp\left[-\frac{\|\bz_{1}^*\|_{2}^{2}}{2\omega^2}-m^2K\log\frac{1}{\epsilon_0}\right]
\end{align}

Since $\|\bz_{1}^*\|_{2}^{2}=O(m^2K)$, we have $\log \Pi(E_2)\geq -m^2K\log \frac{1}{\epsilon_0}$.

%\textbf{Step 5: Rate Derivation}

The convergence rate $\epsilon_0=L^{\frac{1}{3}}m^{\frac{1}{3}}T^{-\frac{1}{3}}n^{-\frac{2}{3}}K^{-\frac{1}{6}}+\sqrt{\frac{m^2 \log{(n^2T/m^2)}}{n^2T}}$ satisfies the required condition:
\begin{equation*}
n^2KT\epsilon_0^2\geq \max\left\{\frac{Lm\sqrt{K}}{\epsilon_0},m^2K\log\frac{1}{\epsilon_0}\right\}
\end{equation*}

This satisfies the condition required by Lemma \ref{lemma:postcontraction}. Applying the lemma with $\alpha = 1/2$ and noting that the Hellinger distance $d'(P_{\calZ}, P_{\calZ^*})$ is bounded by the $\alpha$-divergence $D_\alpha(\bZ, \bZ^*)$, we conclude that:
\begin{equation*}
P\left(\frac{1}{n^2KT}\sum_{t=1}^{T} d'(P_{\calZ}, P_{\calZ^*})\leq M \epsilon^2_{n,m,K,T}\right) \rightarrow 1
\end{equation*}
as $n^2KT \rightarrow \infty$, where $M$ is a sufficiently large constant. This establishes the desired posterior convergence rate for Theorem \ref{the:convergence}.

\section{Detailed Derivations for Model Estimation}

\subsection{E-step: Variational Inference Details}
\label{appx:e-step-details}

In this section, we provide the complete derivations for the E-step optimization. Given the Markovian structure of TSSDMN, the complete data log-likelihood has the form:
\begin{equation}
\begin{split}
     &\log p(\bX, \bZ|\Theta) \\
     & = \log p(\bz_0|\Theta)+\sum_{t=1}^T \log p(\bx_t|\bz_t, \Theta) + \sum_{t=1}^{T}\log p(\bz_t|\bz_{t-1},\Theta) \\
  &= - \frac{1}{2\omega^2}(\bz_0 - \bu_0)^T(\bz_0 - \bu_0) - \frac{1}{2\sigma^2} \sum_{t=1}^T (\bz_t - \bA \bz_{t-1})^T(\bz_t - \bA \bz_{t-1}) \\
& \quad + \sum_{t=1}^T \sum_{j=1}^{n^2K} \log \left(\frac{\exp \bx_{t,j}\bgamma_{t,j}}{1 + \exp \bgamma_{t,j}}\right) - \frac{1}{2} m^2K \log \omega^2 \\
& \quad - \frac{1}{2} Tm^2K \log \sigma^2 + \text{const}.
\end{split}
\end{equation}

The detailed forms of $Q_n^V$ and ELBO are provided in the main text. Here we focus on the gradient computations for the blocked coordinate descent optimization.

The optimization is performed using blocked coordinate descent. The gradients for updating the variational parameters are:

\resizebox{1.03\columnwidth}{!}
{$
\begin{aligned}
   \frac{\partial \underline{{\rm ELBO}}}{\partial \tsigma_t^2} &= \frac{m^2K}{2\tsigma_t^2} - \frac{1}{2} \sum_{j=1}^{n^2K} \frac{\bC^{(v)}_j \bC^{(v) T}_j \exp\left((\bC^{(v)}\tbmu_t + \bb^{(v)})_j + \frac{\tsigma^2_t \bC^{(v)}_j \bC^{(v) T}_j}{2}\right)}{1 + \exp\left((\bC^{(v)}\tbmu_t + \bb^{(v)})_j + \frac{\tsigma^2_t \bC^{(v)}_j \bC^{(v) T}_j}{2}\right)}, \\
   \frac{\partial \underline{\rm ELBO}}{\partial \tbmu_{t,i}} &= - \left(\frac{1}{\sigma^{2(v)}} (\tbmu_t - \bA^{(v)} \tbmu_{t-1} - \bA^{(v) T} (\tbmu_{t+1} - \bA^{(v)} \tbmu_t))\right)_i \nonumber \\
    &\quad - \sum_{j=1} ^ {n^2K}  \frac{\bC_{j,i}^{(v)}\exp\left((\bC^{(v)}\tbmu_t + \bb^{(v)})_j + \frac{\tsigma_t ^ 2 \bC_j^{(v)} \bC^{(v) T}_j}{2}\right)}{1 + \exp\left((\bC^{(v)}\tbmu_t + \bb^{(v)})_j + \frac{\tsigma_t ^ 2 \bC^{(v)}_j \bC^{(v) T}_j}{2}\right)}  + (\bC^{(v) T} \bx_t)_i.
\end{aligned}$
}

\begin{algorithm}[t]
\caption{Blocked coordinate descent for E-step}\label{alg:bcd}
\begin{algorithmic}
\STATE {\bfseries Input:} The observations $\bX$, the current model parameters $\Theta^{(v)}$, tolerance $h_{\rm tol}$.
\STATE {\bfseries Result:} The parameters in $q(\bZ)$: $\tbmu_t, \tsigma_t^2, \forall t=0,\ldots,T$.
\STATE Initialize $\tbmu_t^{(0)}$ and $\tsigma_t^{2(0)}$ for all $t = 0, 1, \ldots, T$;
\REPEAT
    \FOR{$t = 0, 1, \ldots, T$}
        \STATE Fix $\tbmu_0^{(s)}, \ldots, \tbmu_{t-1}^{(s)}, \tbmu_{t+1}^{(s-1)}, \ldots, \tbmu_T^{(s-1)}$ and $\tsigma_0^{2 (s)}, \ldots, \tsigma_{t-1}^{2 (s)}, \tsigma_{t+1}^{2 (s-1)}, \ldots, \tsigma_T^{2 (s-1)}$, update $\tbmu_t$ and $\tsigma_t^2$ by gradient descent.
    \ENDFOR
    \STATE $s \leftarrow s + 1$.
\UNTIL{$\sum_{t=0}^T (\tbmu_t^{(s)}-\tbmu_t^{(s-1)})^2+(\tsigma_t^{2\ (s)}-\tsigma_t^{2\ (s-1)})^2 \leq h_{\rm tol}$}
\RETURN $\tbmu_t^{(s)},\tsigma_t^{2(s)},\forall t=0,\ldots,T$.
\end{algorithmic}
\end{algorithm}

\subsection{M-step: Parameter Update Details}
\label{appx:m-step-details}

The objective function for observation parameters is:
\begin{align}
    l_C(\bC_1,\bC_2, \bb) &= \sum_{t=1}^T \bx_t^T (\bC \tbmu_t + \bb) \\ 
    &- \sum_{t=1}^T \sum_{j=1} ^{n^2K} \log\left(1 + \exp \left((\bC \tbmu_t + \bb)_j + \frac{\tsigma^2_t \bC_j\bC_j^T }{2} \right) \right).
\end{align}

The Kronecker product structure requires careful computation of partial derivatives:
\begin{align}
    \frac{\partial \bC}{\partial \bC_{1,xy}} &= \bC_2 \otimes (\bC_1 \otimes \bDelta_{1,xy} + \bDelta_{1,xy} \otimes \bC_1), \\
    \frac{\partial \bC}{\partial \bC_{2,xy}} &= \bDelta_{2,xy} \otimes \bC_1 \otimes \bC_1,
\end{align}
where $\bDelta_{i,xy}$ are indicator matrices with 1 at position $(x,y)$ and 0 elsewhere.

For the bias parameter:
\begin{equation}
    \frac{\partial l_C}{\partial \bb_i} = \sum_{t=1}^T \bx_{t,i} - \sum_{t=1}^T \frac{ \exp\left(\bC_j \tbmu_t + \bb_i + \frac{\tsigma^2_t \bC_i \bC^T_i}{2}\right)}{1 + \exp\left(\bC_j \tbmu_t + \bb_i + \frac{\tsigma^2_t \bC_i \bC^T_i}{2}\right)}.
\end{equation}

For the dynamics parameters $\{\bA_1, \bA_2, \bA_3\}$, we maximize:
\begin{align}
    l_3(\bA_1, \bA_2, \bA_3) &= - \frac{1}{2} \sum_{t=1}^T \| \tbmu_{t}  - \bA \tbmu_{t-1} \|_2 ^2 - \frac{1}{2}  {\rm tr}(\bA \bA^T) \sum_{t=1}^T \tsigma_{t-1}^2.
\end{align}

The gradient with respect to $\bA_{i,xy}$ is:
\begin{align}
    \frac{\partial l_3}{\partial \bA_{i,xy}} &= -\sum_{t=1}^{T} {\rm tr}\left(\left(\frac{\partial{\bA}}{\partial \bA_{i,xy}}\right)^T(\tbmu_t - \bA \tbmu_{t-1}) \tbmu_{t-1}^T\right) \nonumber \\
    &\quad - (T-1) \sum_{t=1}^T \tsigma_t^2 {\rm tr}\left(\bA^T \left(\frac{\partial{\bA}}{\partial \bA_{i,xy}}\right)\right).
\end{align}

The complete gradient expressions and numerical implementation details for all parameters are straightforward applications of the chain rule, considering the Kronecker product structure in the derivatives.

\begin{algorithm}[t]
    \caption{ EM algorithm for TSSDMN}\label{alg:EM}
    \begin{algorithmic}
    \STATE{\bfseries Input:} The observations $\bX$, the maximum iteration $v_{\max}$.
    \STATE{\bfseries Results:} The parameters of dynamic multilayer network $\hat{\Theta} = \{ \hat{\bC}_1, \hat{\bC}_2, \hat{\bA}_1, \hat{\bA}_2, \hat{\bA}_3, \hat{\bb},\hat{\bu}_0,\hat{\sigma}^2,\hat{\omega}^{2}\}$;
    \STATE Initialized parameter $\Theta^{(0)}, v \leftarrow 0$; 
    \WHILE{$v \leq v_{\max}$}
    \STATE Obtain the posterior distribution $q(\bZ)$ under parameter $\Theta^{(v)}$ by Algorithm \ref{alg:bcd}; 
    \STATE Obtain $\bC^{(v+1)}_1, \bC^{(v+1)}_2, \bb^{(v+1)}$ by projected gradient descent or gradient descent methods. % via Eqs. (\ref{equ:projected gradient descent}), (\ref{eq:C2_derivative}), (\ref{eq:b_derivative}); 
    \STATE Obtain $\bA^{(v+1)}_1, \bA^{(v+1)}_2, \bA^{(v+1)}_3$ by gradient descent method %.via Eq. (\ref{eq:A_derivative}); 
    \STATE Obtain $\bu_0^{(v+1)},\omega^{2(v+1)},\sigma^{2(v+1)}$ by Eqs. (\ref{equ:update-u}), (\ref{equ:update-omega}), (\ref{equ:update-sigma});
    \STATE $v \leftarrow v + 1$; 
    \ENDWHILE
    \RETURN $\hat{\Theta} = \Theta^{(v_{\max})}$
\end{algorithmic}
\end{algorithm}

\end{document}